\documentclass[twocolumn,fleqn]{svjour2}
\smartqed  % flush right qed marks, e.g. at end of proof
\newcommand{\etal}{\textit{et al.}}
\usepackage{algorithm}
\usepackage{algpseudocode}
\usepackage{graphicx}
\usepackage{mathtools}
\journalname{arXive}
\begin{document}

\title{Real-time Pedestrian Surveillance with Top View Cumulative Grids%\thanks{Grants or other notes
}


\author{Kai Berger         \and
        Jeyarajan Thiyagalingam 
}


\institute{K. Berger \at
              OerC Oxford \\
              \email{kai.berger@oerc.ox.ac.uk}           %  \\
           \and
           J. Thiyagalingam \at
              OerC Oxford\\
 \email{jeyarajan.thiyagalingam@oerc.ox.ac.uk} 
}

\date{Received: date / Revised: date}

\maketitle

\begin{abstract}
This manuscript presents an efficient approach to map pedestrian surveillance footage to an aerial view for global assessment of features. 
The analysis of the footages relies on low level computer vision and enable real-time surveillance. While we neglect object tracking, we introduce cumulative grids on top view scene flow visualization to highlight situations of interest in the footage. Our approach is tested on multiview footage both from RGB cameras and, for the first time in the field, on RGB-D-sensors.
\end{abstract}

\section{Introduction}
\label{intro}
%Your text comes here. Separate text sections with
%\section{Section title}
%\label{sec:1}
%Citation of \citet{RefJ}.
%\subsection{Subsection title}
%\label{sec:2}
%as required. Don't forget to give each section
%and subsection a unique label (see Sect.~\ref{sec:1}).
%\paragraph{Paragraph headings} Use paragraph headings as needed.
%\begin{equation}
%\end{equation}

The security of public areas is a critical part for pluralistic society.
While we enjoy the freedom to use public facilities ranging from bus stations to airports for personal purposes,
we have to be aware that these places are also the most likely areas for an impending attack with
a high death toll. Apart from terroristic attacks, burglary and beggary are also most likely occur there, because these places
bear with a great amount of anonymity for the subjects using them. Therefore, authorities strive to
increase security with public surveillance (CCTV). Apart from the psychological factor, the camera system
also directly helps avoiding crimes by live streams which help monitoring officers to alert ambulance. Further,
recorded footage might later help convict suspects once the charges have been pressed by victims.
Human operation to CCTV monitoring however introduces the risks that crucial events are missed due to attention shifts
or blind spots in the surveillance area. Furthermore, when the video footage is juxtaposed on monitors, it becomes a 
non-trivial task for the monitoring officer to build a mental map from the image streams. A subject leaving the imaged
area in one window might not appear afterwards in neighboring windows. To tackle this problem, several high-level
computer vision algorithms have been proposed, to recognise, label and track subjects image by CCTV. However, a failure
of such a system to detect a single subject once can become fatal, if that subject turns out to be the villain, or a terrorist.
\begin{figure}
\centering
  \includegraphics[height=0.11\textwidth]{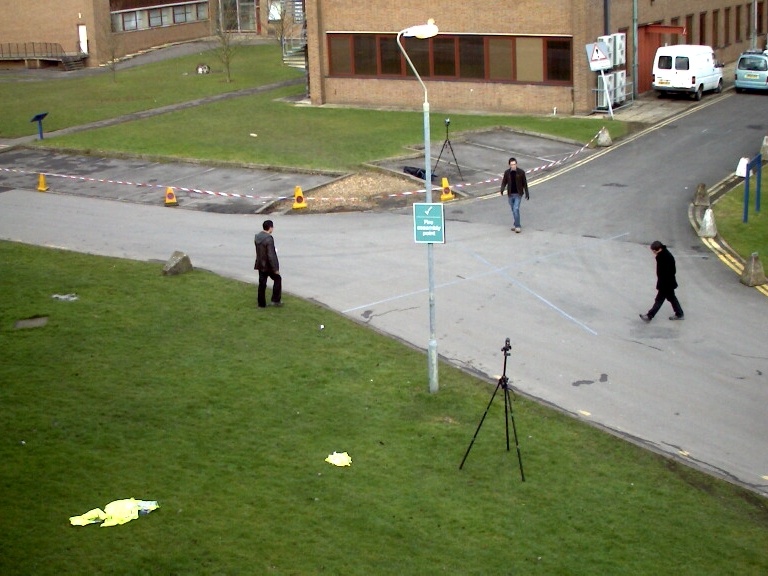}
\includegraphics[height=0.11\textwidth]{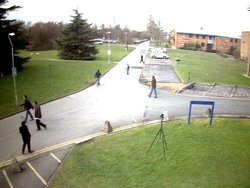}
  \includegraphics[height=0.11\textwidth]{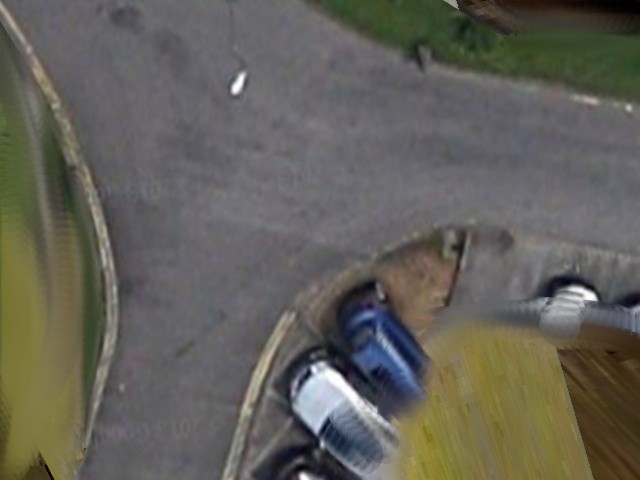}

\caption{Input footage from a calibrated viewpoint (left, middle) and simplified satellite map (right).}
\label{fig:1}       % Give a unique label
\end{figure}
In this paper we want to overcome the weaknesses of both the juxtaposition of image streams for human-based monitoring
and the high-level vision based automatic tracking. We propose to build the mental map virtually by rendering a top view
from one to many input cameras. The evaluation of of the depicted motion is left to the human monitoring officer but our algorithm
visualizes occupancy times and the scene flow on the top view to help with the judgement of a given situation.
For the first time, we present an approach that incorporates traditional RGB camera based scene surverillance and
depth-value based scene surveillances, e.g. from RGB-D sensors or Time-Of-Flight cameras.
This enables new hardware setups for public areas, based on depth-sensors that have increasingly gotten affordable over the last three years
and that help surveying scenes with varying lighting conditions.

The paper is structured as follows: after briefly revising the state of the art in Sect.~\ref{relWork}, we will present an approach to retarget pedestrian surveillance footage into the topview in order to perform motion and video analysis to it, Sect.~\ref{proposed}.
Two algorithms, one for RGB-data and one for depth data will described in detail. The motion analysis consists of cumulative maps to help crowding detection
and the application of optical flows. In Sect.~\ref{results}, we will apply the algorithm to an RGB outdoor dataset consting of eight calibrated cameras and two indoor datasets, before we conclude in Sect.~\ref{conclusion}. One dataset is a depth-based dataset consisting of one ToF-camera, the other is the surveillance of a train station with for extrinsically calibrated RGB cameras.
\section{Related Work}
\label{relWork}
\textbf{Reprojection and summarization}

Image reprojecyion aims to warp an input image into an arbitrary viewpoint and a typical implementation using an energy functional minimization is formulated by Setlur \etal \cite{setlur2005automatic}.
Carroll \etal~\cite{Carroll:2010:IWA:1778765.1778864} reprojects images in order to artistically introduce a new perspective to a given image, while 
  Sacht \etal~\cite{sacht2011scalable} reprojects photographic images into specific cylindrical views.
The term \textit{summarization} was coined by Daniel and Chen \cite{vidvis} to reproject an image sequence meaningfully into one single image and the was later refined by Botchen \etal \cite{Botchen:2008:ABM}. Based on this approach it was shown \cite{vis06-chen} that ordinary users can learn to detect and recognise \emph{visual signatures} of events from video visualization. 
Wang \etal\ \cite{Wang:2007:TVCG} proposed to reproject videos into a 3D environment model for scene awareness.  Legg \etal used homographic projection for 3D reconstruction from a single viewpoint \cite{Legg:2011:ICIP} and Parry \etal applied this approach to image sequences in sports~\cite{Parry:2011:TVCG}. Remero \etal\ \cite{Romero:2008:TVCG} summarized activities captured by an aerial camera in natural settings.
Further information can be found in a comprehensive survey  on video-based graphics and video visualization conducted by Borgo \etal~\cite{Borgo:2013:CGF}.
\begin{figure*}
\centering
\includegraphics[width=0.15\textwidth]{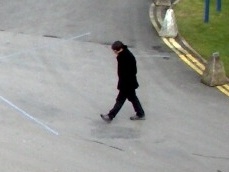}
\includegraphics[width=0.15\textwidth]{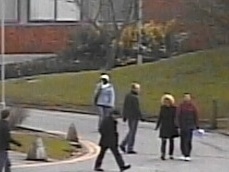}
  \includegraphics[width=0.15\textwidth]{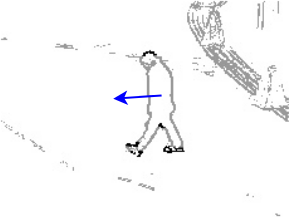}
  \includegraphics[width=0.15\textwidth]{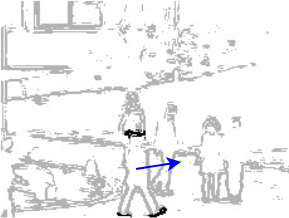}
\includegraphics[width=0.15\textwidth]{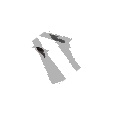}
\includegraphics[width=0.15\textwidth]{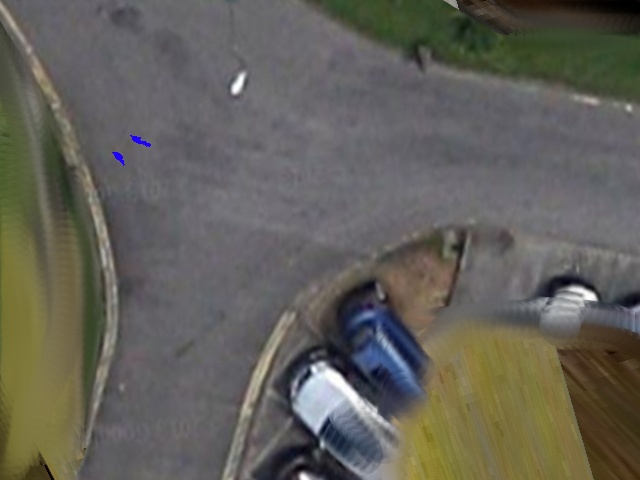}
\caption{A pedestrian captured by two cameras (first, second) is mapped to a top-view by finding edge segments colinear with the mean optical flow vector of the pedestrian in each camera image (third, fourth). Corresponding areas in between the edge segments are intersected (fifth) to get the overlay in the aerial view (sixth).}
\label{fig:2}       % Give a unique label
\end{figure*}

\textbf{Pedestrian Surveillance}
Classically computer vision strives to detect pedestrians in input footage rather than reprojecting them meaningfully.
For example Bayesian classifiers \cite{article25} to evaluate the presence of an object are widely employed for detecting pedestrians \cite{article11,article12}.
Multi-person tracking has been implemented differently: the authors of \cite{article22} track pedestrians by flow optimization on the ground plane. The authors expand the algorithm with global appearance rules. The method proposed in \cite{article3} labels pedestrians the problem by estimating both discrete data and continuous trajectories using global costs. 
This method relies solely on trajectories and does not involve appearance of
 Generally, association between data can be performed over a list of frames \cite{article15}, e.g. in a hierarchical manner.
In \cite{article17} the authors use tracklets from lower levels in the pyramid and associate them accordingly at a higher level to reduce the computational load.
This method introduced in \cite{article17} is hierarchical, for example, and uses existing low-level tracklets.
 In \cite{article14},  the tracks of pedestrians are refined with motion models. 
Multi-view methods have
been introduced to overcome occlusion introduced in crowding scenarios.
Currently, some multi-view based papers \cite{article20,article21} learn context models. 
Cowd density estimations can be used \cite{article19} to improve human detection and tracking. 
The authors of \cite{article21} propose an adaptation scheme in which classifiers
are learned incrementally through online boosting, so that they adapt to the changes over time. In \cite{article4}, the authors propose
to match a bipartite graph.

\textbf{RGB-D Imaging}
Shotton et al.~\cite{shotton2011real} introduced the Kinect and its underlying algorithm as a tool to capture the human pose from monocular depth images and paved the road for consumer-grade
motion capturing with the device. In the wake of the commercial success monocular motion capturing has become the focus of the research community~\cite{girshick2011efficient,raptis2011real,nowozin2011decision}. Mainly, the Microsoft Kinect was used capture datasets and benchmarks. Besides the tracking of limbs and joints quickly other research fields in monocular depth processing have emerged.

One research direction for example was to use the Microsoft Kinect as a hand-tracking device ~\cite{oikonomidis2011efficient}. Frati \etal~\cite{frati2011using} assume the hand to always be closest limb to the camera  while Reheja and his colleagues detect the palm with a circular filter the depth image~\cite{raheja2011tracking}. Van den Bergh et al.~\cite{van2011real} estimate the orientation of the hand from the orientation of the forearm in the depth image. Zollhofer et al.~\cite{zollhofer2011automatic} fitted deformable facial meshes to depth data captured
from human faces by relying on feature points (eyes, nose) in the depth data. Leyvand et al. also examine the face recognition of identical twins given depth and motion data from the Microsoft Kinect~\cite{leyvand2011kinect}. 

In 2011, Berger and his colleagues showed, that it is also possible to employ multiple depth sensors in one scene for motion capturing research~\cite{berger2011markerless}. Using an external hardware shutter~\cite{schroder2011multiple} they were able to reduce the sensor noise introduced from neighboring Kinects. A similar approach has been introduced by Maimone and Fuchs~\cite{maimone2012reducing}: each Kinect sensor shakes around its up vector introducing scene motion. This approach was refined by Butler \etal~\cite{butler2012shake} who introduce arbitrary motion to the sensor which has to hang in an acryllic frame and has to be held by rubber bands.

Beside the Kinect sensor, depth imaging is usually conducted by passive stereo or with Time of Flight imaging.
Most methods use the range imaging data to initialize stereo matching and impose constraints on the search range depending on the range imaging and stereo noise model.
Local methods \cite{kuhnert2006fusion,gudmundsson2008fusion,hahne2009depth,DalMutto3DPVT10,mmsp-10-qingxiong-yang,nair2012high} combine the stereo and the range imaging data 
term on a per pixel level.
Kuhnert et al. and Hahne et al.\cite{kuhnert2006fusion,hahne2009depth} compute confidences in the depth image and let stereo refine the result in regions with low confidence.
Nair et al.\cite{nair2012high} and Dal Mutto et al. \cite{DalMutto3DPVT10} combine confidences from both stereo and the depth image into a stereo matching framework.
Global methods such as \cite{fischer2011combination,hahne2008combining,kim2009multi,nair2012high} use spatial regularization techniques in order to 
propagate more information to regions with low stereo or depth image confidence.
Among the global methods, Fischer et al.\cite{fischer2011combination} apply an extension of half global matching \cite{hirschmueller2008} to additionally handle depth data.
Hahne et al.\cite{hahne2008combining} applied a graph cut approach with a discrete number of disparities to sensor fusion. 
Kim et al. \cite{kim2009multi}  and Nair et. al \cite{nair2012high} formulate the fusion problem in as a energy functional that is then minimized.
Nair et al.\cite{nair2012high} employ adaptive first and second order total variation (TV) with L1 regularization. normally used to estimate optical flow. 

\section{Proposed Algorithm and Test Datasets}
\label{proposed}

In this section we describe the algorithm for top view summarization maps in detail. Its key idea is outlined in Fig. \ref{fig:2}. For each input view $v \in V_{in}$ the algorithm, Alg. \ref{algretarget}, requires a background image $I_{v,bg}$ and two consecutive images $I_{v,t},I_{v,t-1}$. A background image can either be manually set by the user ($I_{v,bg} = I_{v,t_{\text{user}}}$) or aquired as mean intensity image over a time span $T$ ($I_{v,bg} = \frac{\sum_t I_{v,t}}{T}$). In order to reproject pixel areas comprised by pedestrians into a topview with a calibrated Homography $W_v: v \in V_{in} \rightarrow v_{\text{topview}} \in V_{\text{target}}$ the algorithm searches for regions near edges that move parallel to the mean direction of motion for each pedestrian. This can be reasoned by the fact, that in a head-and-shoulder view, i.e. a top view $v_{\text{topview}} \in V_{\text{target}}$, only head, should and feet are visible and these body parts mainly contribute to edges $e_{\parallel} \in I_{\text{edge},v,t}$ in an edge image $\in I_{\text{edge},v,t}$ that are parallel to the mean direction of motion $\vec{v}_{\text{pedestrian}}$ of a pedestrian in a corresponding flow image $I_{\text{flow},v,t}$, while most other body parts appear perpendicular. Thus, the algorithm computes the edge image $I_{\text{edge},v,t}$ of the latest frame and, using the previous frame, it computes the mean flow vector for each pedestrian, i.e. each connected component $CC_{v,t}$ using the optical flow evaluation $I_{\text{flow},v,t}$.
This is done by calling Alg. \ref{algof}. The computation of connected components is performed using the background image $I_{v,bg}$ and a significant colour distance. After having retrieved the mean optical flow vector for each connected component, which is equal to the mean direction of motion for each pedestrian in the image, Alg. \ref{algretarget} calls Alg. \ref{algat}, which thresholds each edge segment for its collinearity to the mean flow of each object respectively. The threshold is performed by computing the angular difference between the flow vector and the normal to the gradient at each edge pixel and thresholding for the angular difference. Pixel regions in between the remaining edge pixels are filled in a line-sweep manner to arrive at a subset of the connected component pixels. Using the camera calibration, Alg. \ref{algretarget} retargets the remaining pixels into the top view. To avoid distortions that linearly increase with distance form the ground plane, the algorithm intersetcs the reprojected pixel sets from all viewpoints to arrive a the intersection set coinciding with the area actually comprised by the pedestrian in a hypothetical top view.
In table~\ref{tab:1} we show the outcome of Alg. \ref{algretarget} on the PETS 2007 database~\cite{pets09}. In that scenario cameras are installed on the campus located at $51°26'18.5N$ $000°56'40.00W$ to cover an approximate area of 100m x 30m. The frames from different views can be considered as synchronised, but there are slight delays and frame drops detectable on rare occasions. The footage was not colour-corrected afterwards. The calibration has been performed with visual markes and Tsai Camera Calibration. The ground plane is assumed to be the Z=0 plane. All spatial measurements have been conducted in metres. The recorded frames were compressed as JPEG image sequences. Different sequences depict random walking crowd flow and regular walking pace crowd flow of different sparsity and with varying lighting conditions ranging from overcast to bright sunshine. Further walking, running, evacuation, i.e., rapid dispersion, local dispersion, crowd formation and splitting at different time instances have been simulated by the actors. Note, that we apply the motion analysis to the sequence in the topview, by computing a cumulative grid, Table \ref{tab:1}  third row, and the optical flow for each pixel set, Table \ref{tab:1}  fourth row. The cumulative grid is implemented by applying a cumulative moving average $A_t=\frac{p_1(x,y)+ \cdots + p_t(x,y)}{t}$. We may constrict the time span of the cumulative grid to a reasonable value $t_{span}$, for example several minutes. The implementation may then be altered to arrive at the following formula $A_t=A_{t-1}-\frac{p_{t-t_{span}}(x,y)}{t_{span}}+\frac{p_t(x,y)}{t_{span}}$.
\begin{table*}[t]
\caption{Analysis and summarization of a pedestrian database~\cite{pets09}. The crowding that is noticable at time frames $t+2,t+3,t+4$ in the input images (first row) can be directly read from the cumulative top view scatterplot (third row). An optical flow computation can be applied to the top view, e.g. to discriminate pixel movements or search for unusual motion patterns.}
\centering
\label{tab:1}       % Give a unique label
\begin{tabular}{cccccccc}
\hline\noalign{\smallskip}
\includegraphics[width=0.1\linewidth]{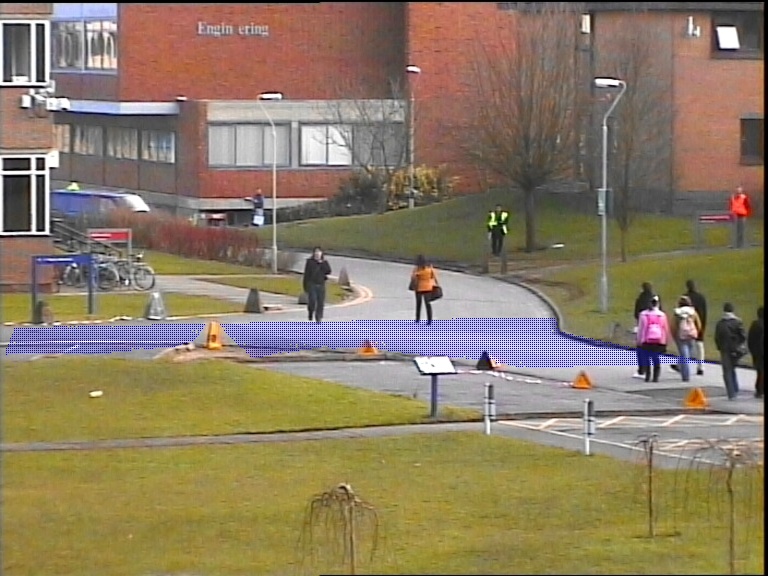} & \includegraphics[width=0.1\linewidth]{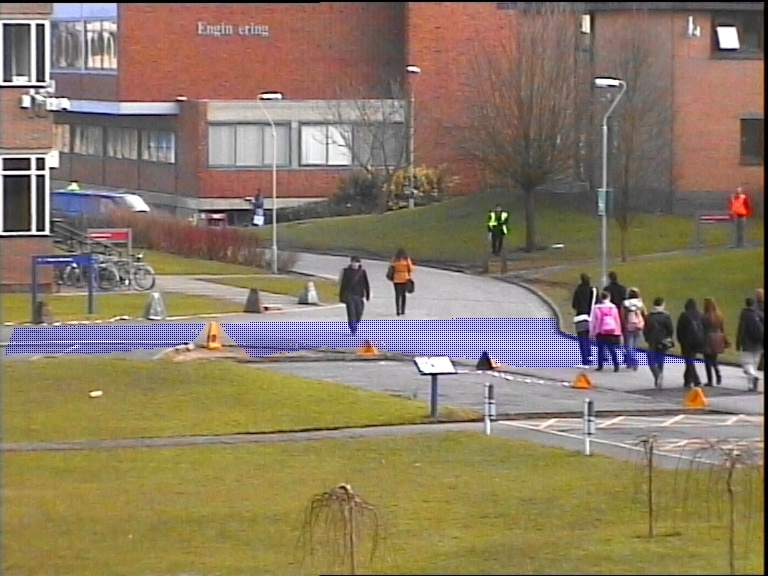} &\includegraphics[width=0.1\linewidth]{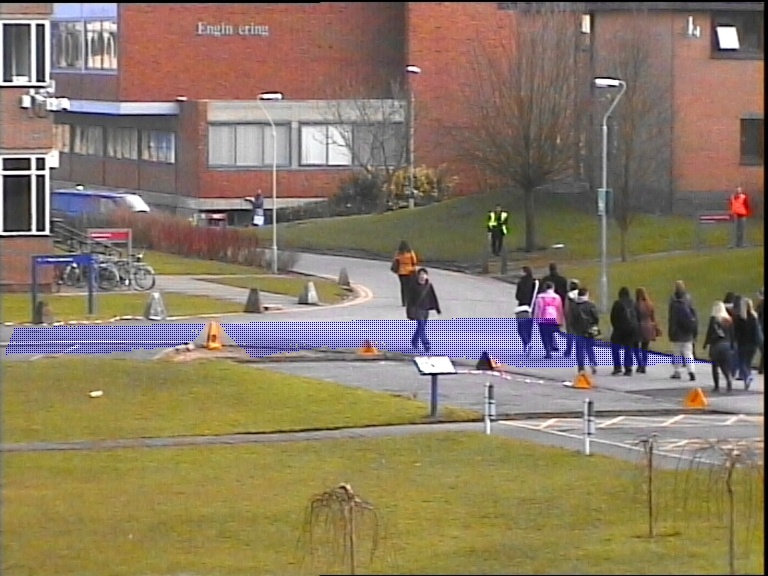} &\includegraphics[width=0.1\linewidth]{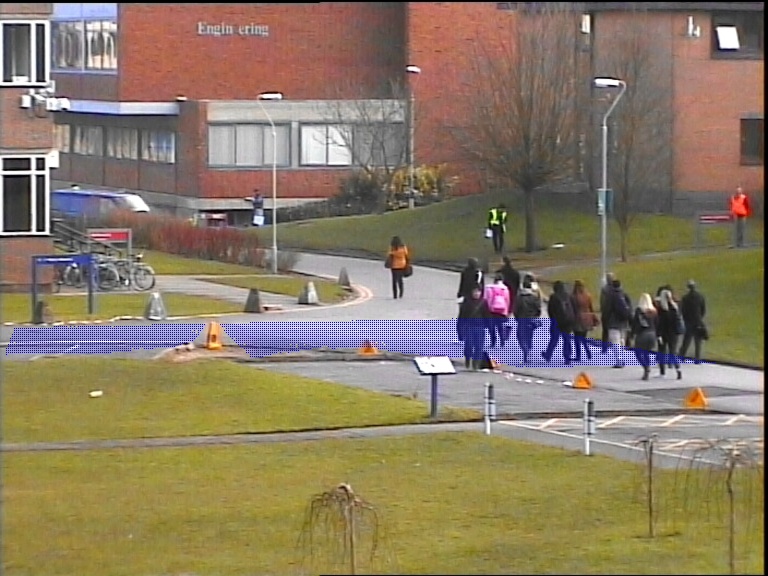} &\includegraphics[width=0.1\linewidth]{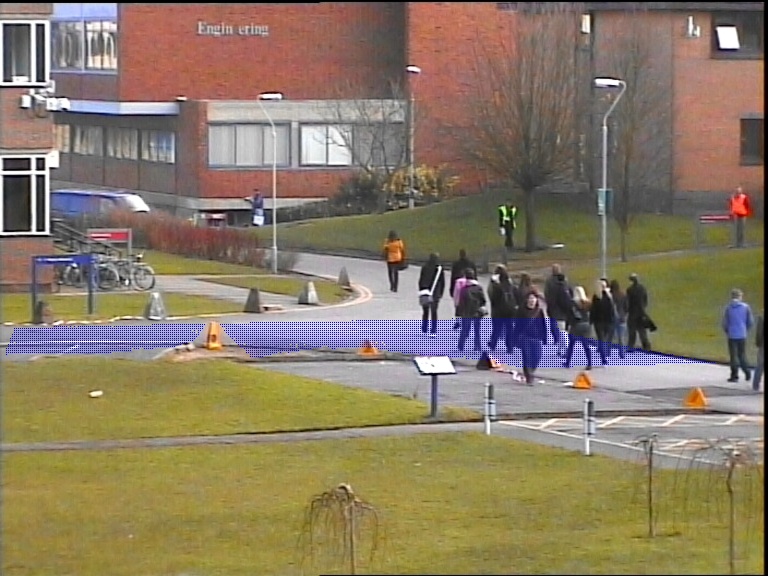}&\includegraphics[width=0.1\linewidth]{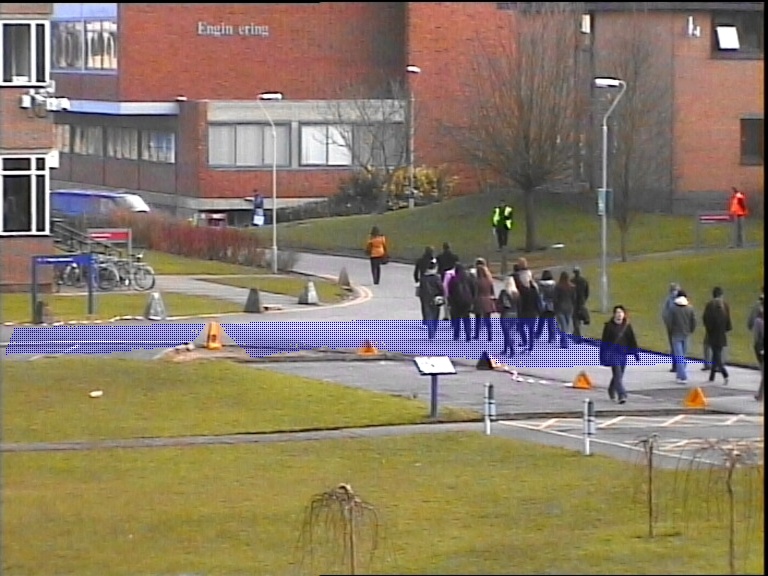}&\includegraphics[width=0.1\linewidth]{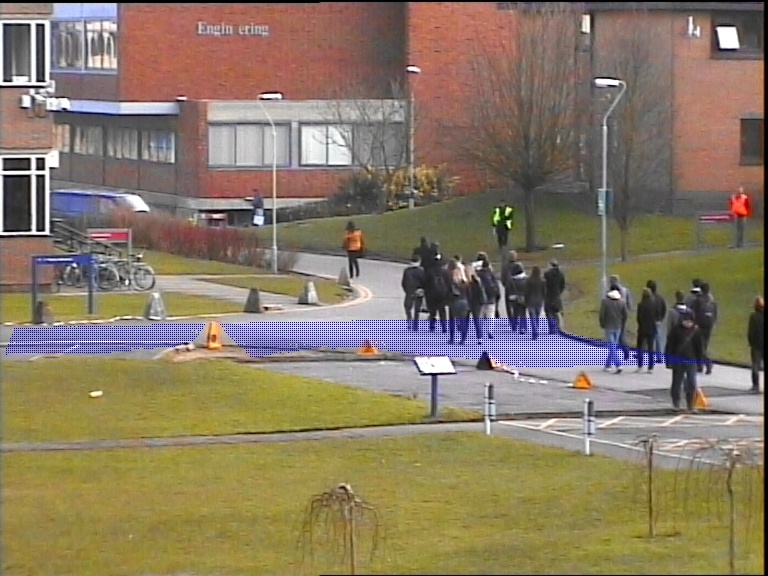}&\includegraphics[width=0.1\linewidth]{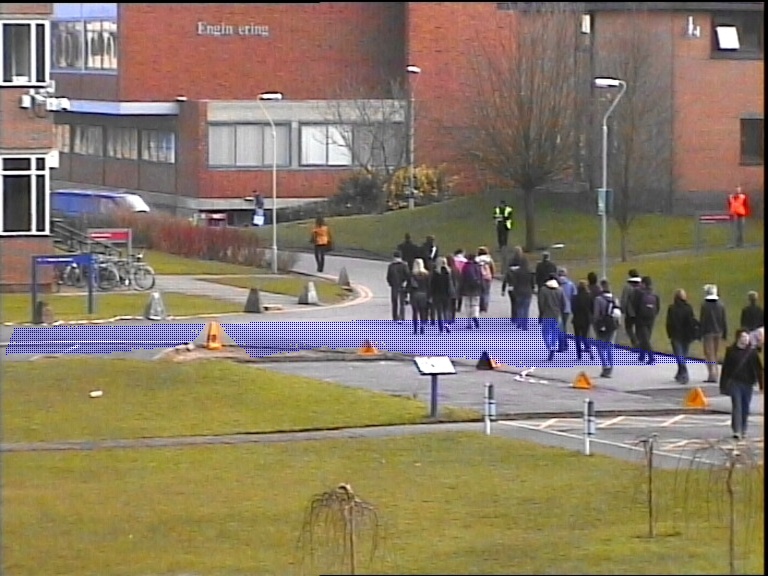}\\ \hline\noalign{\smallskip}
$t$&$t+1$&$t+2$&$t+3$&$t+4$&$t+5$&$t+6$&$t+7$\\
\\ \hline\noalign{\smallskip}
\\
\includegraphics[width=0.1\linewidth]{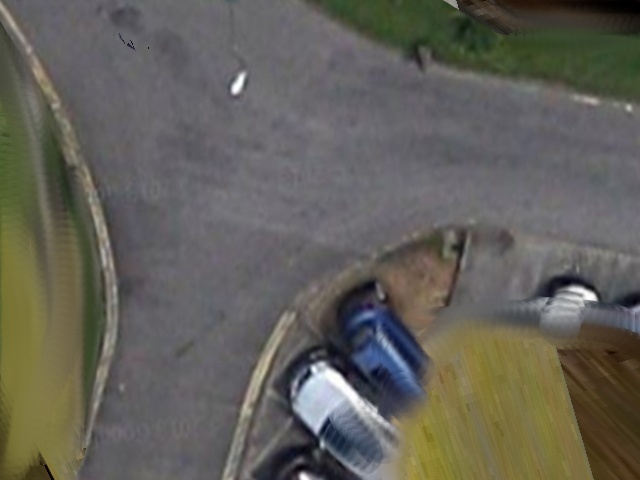} & \includegraphics[width=0.1\linewidth]{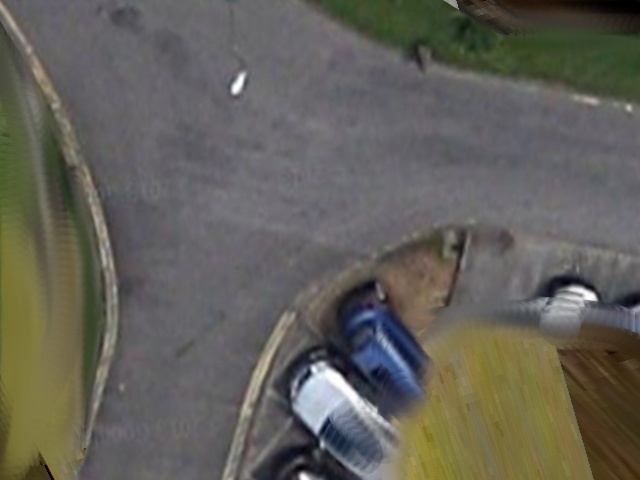} &\includegraphics[width=0.1\linewidth]{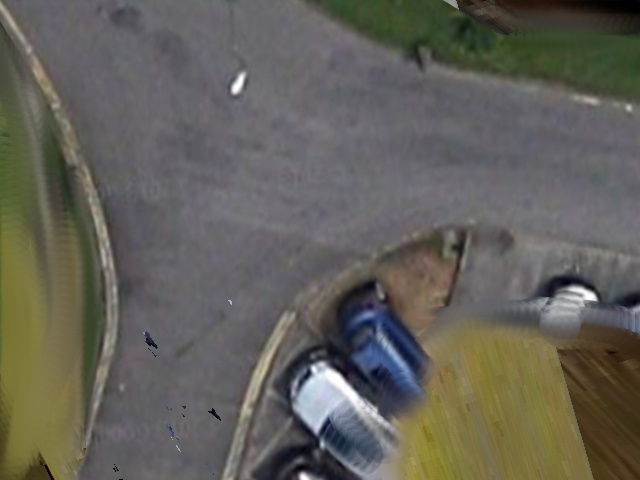} &\includegraphics[width=0.1\linewidth]{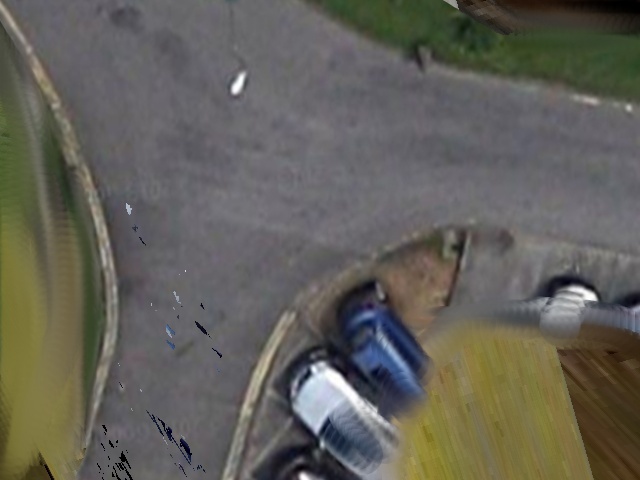} &\includegraphics[width=0.1\linewidth]{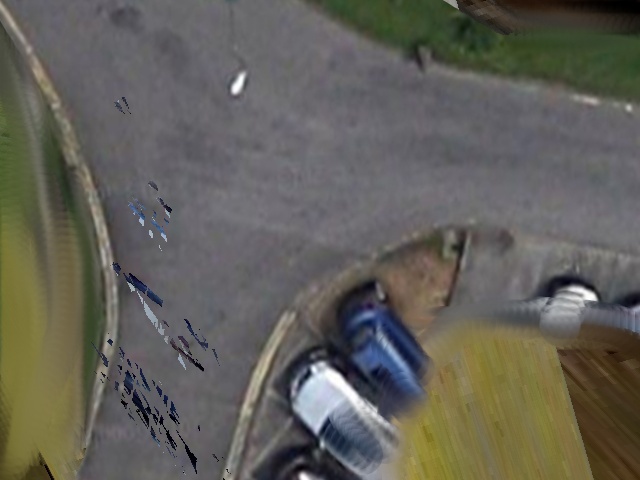}&\includegraphics[width=0.1\linewidth]{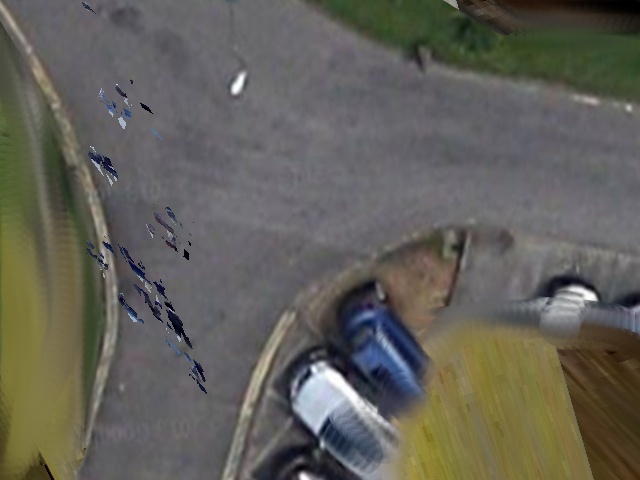}&\includegraphics[width=0.1\linewidth]{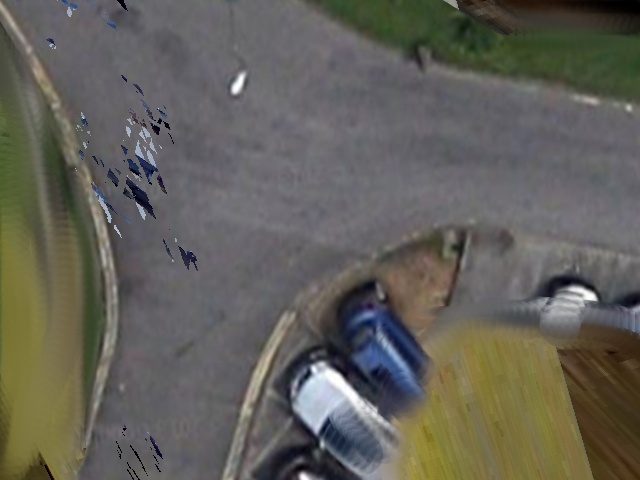}&\includegraphics[width=0.1\linewidth]{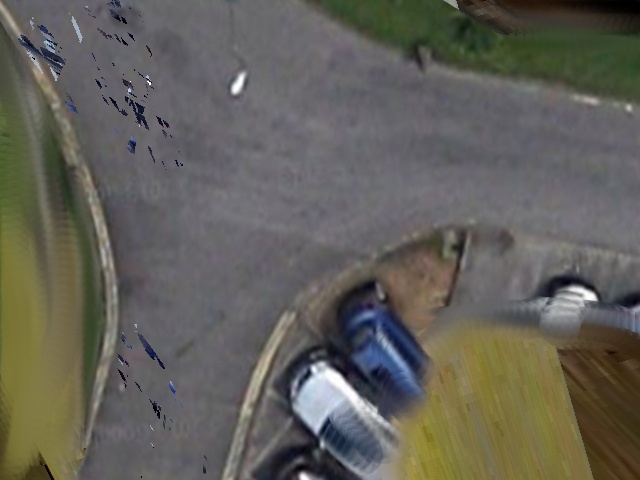} \\
\includegraphics[width=0.1\linewidth]{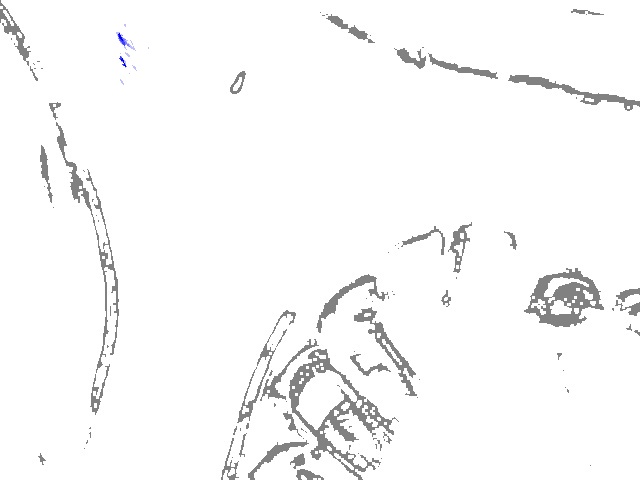} & \includegraphics[width=0.1\linewidth]{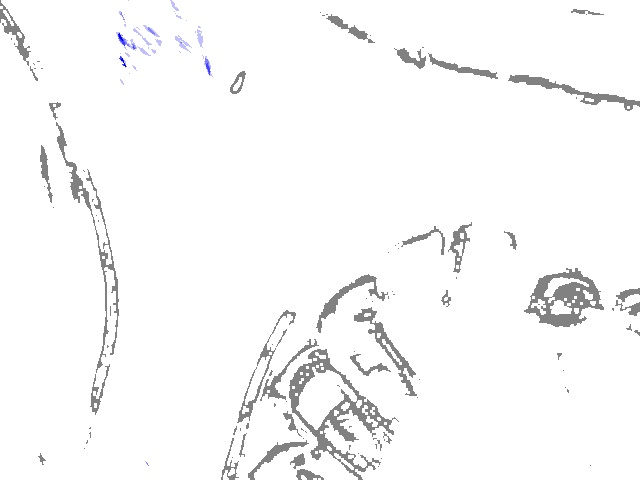} &\includegraphics[width=0.1\linewidth]{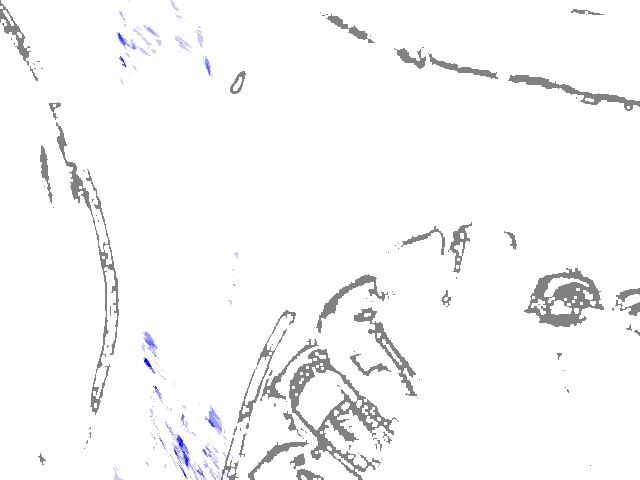} &\includegraphics[width=0.1\linewidth]{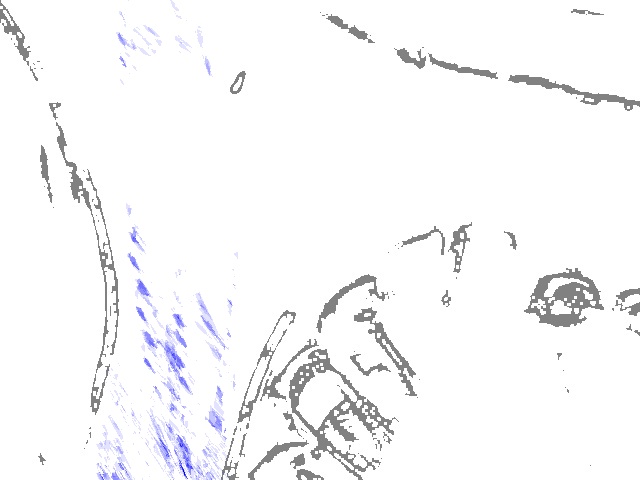} &\includegraphics[width=0.1\linewidth]{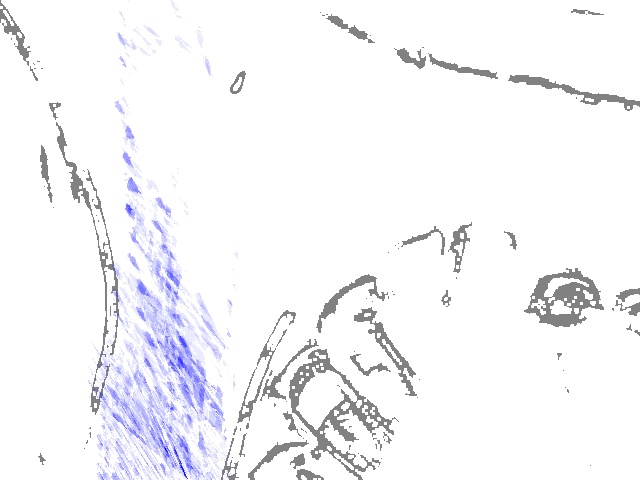}&\includegraphics[width=0.1\linewidth]{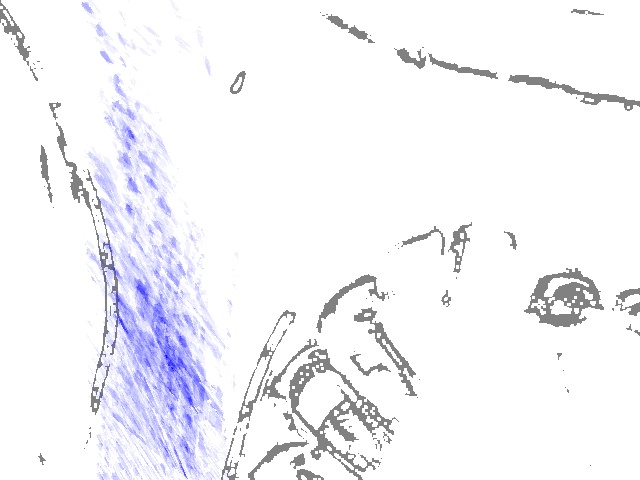}&\includegraphics[width=0.1\linewidth]{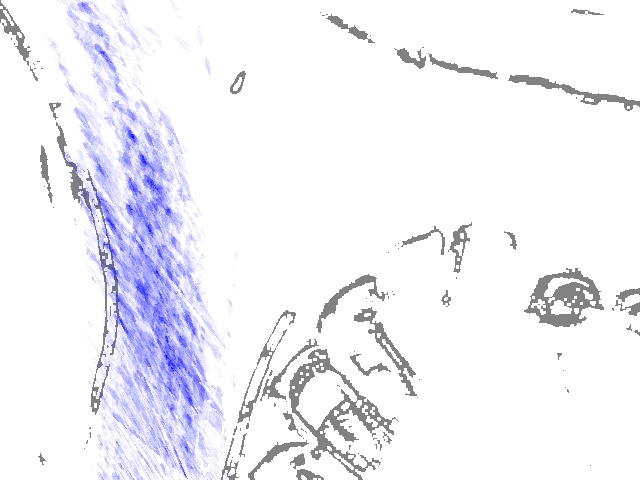}&\includegraphics[width=0.1\linewidth]{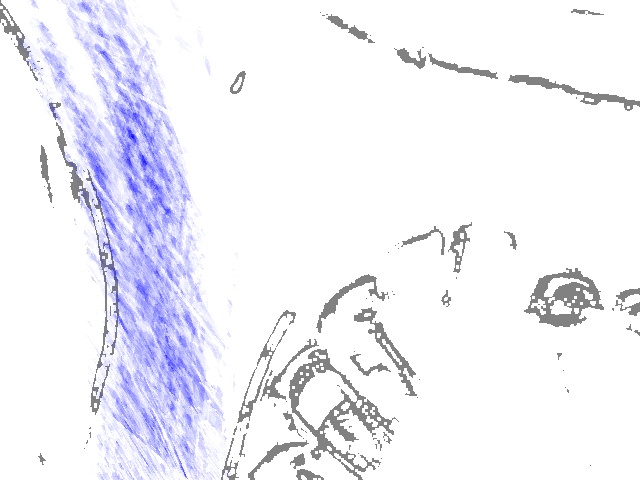} \\
\includegraphics[width=0.1\linewidth]{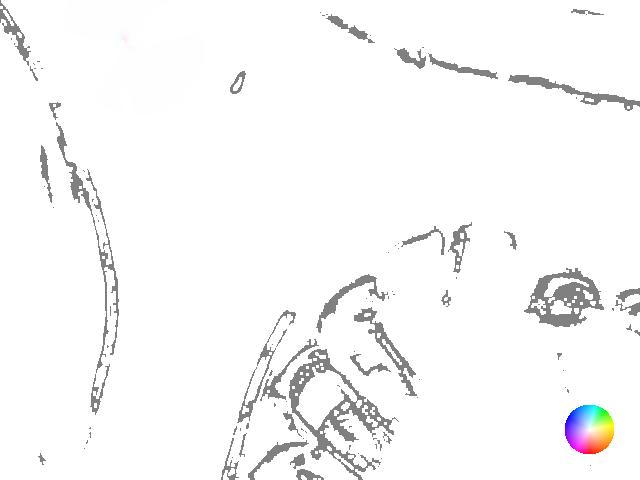} & \includegraphics[width=0.1\linewidth]{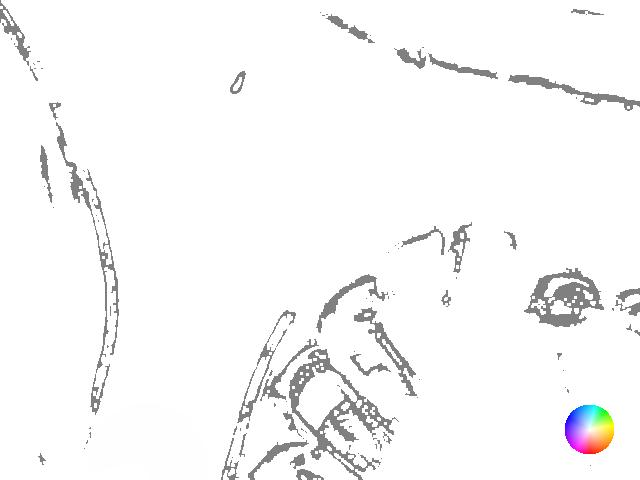} &\includegraphics[width=0.1\linewidth]{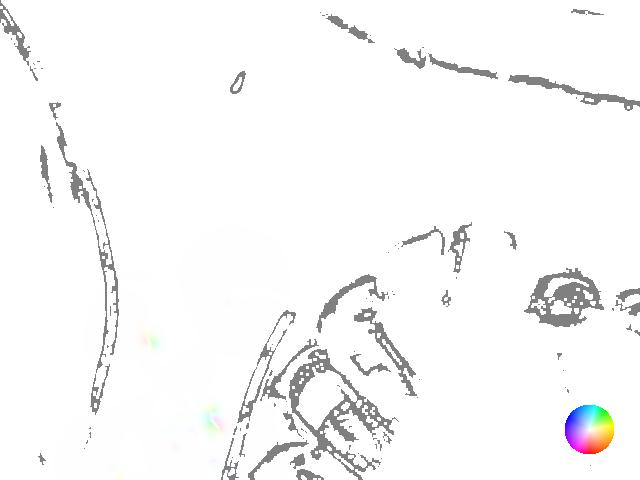} &\includegraphics[width=0.1\linewidth]{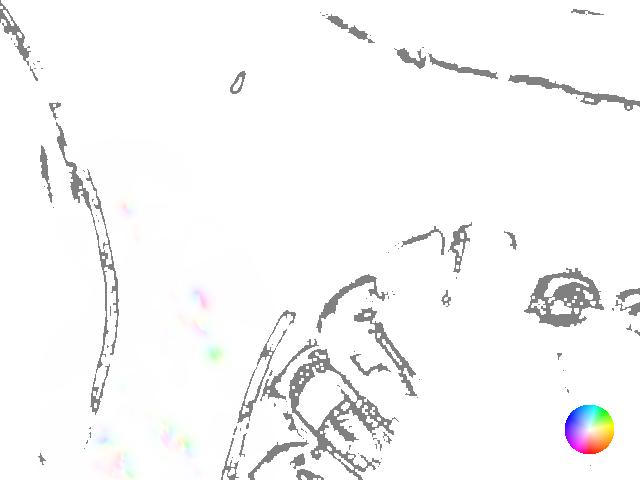} &\includegraphics[width=0.1\linewidth]{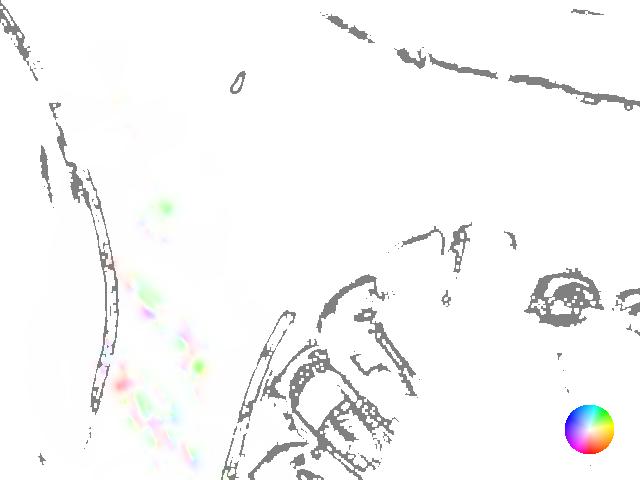} &\includegraphics[width=0.1\linewidth]{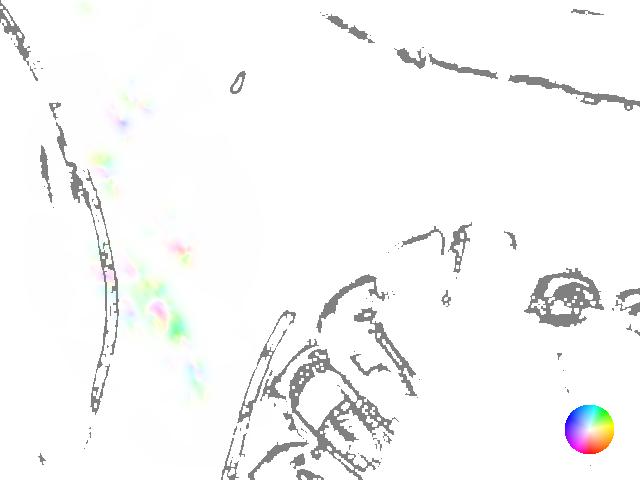}&\includegraphics[width=0.1\linewidth]{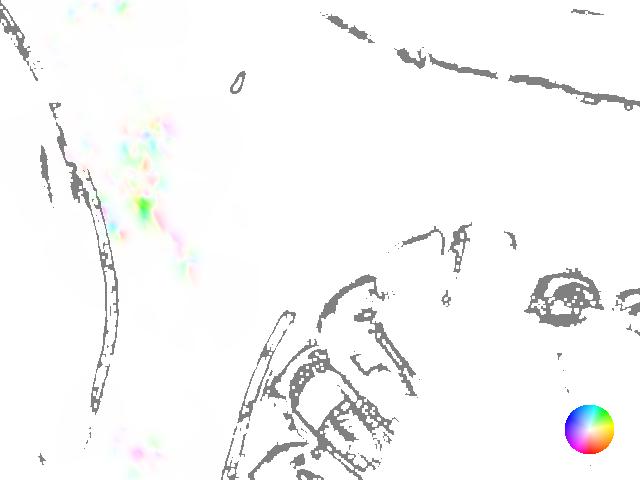}&\includegraphics[width=0.1\linewidth]{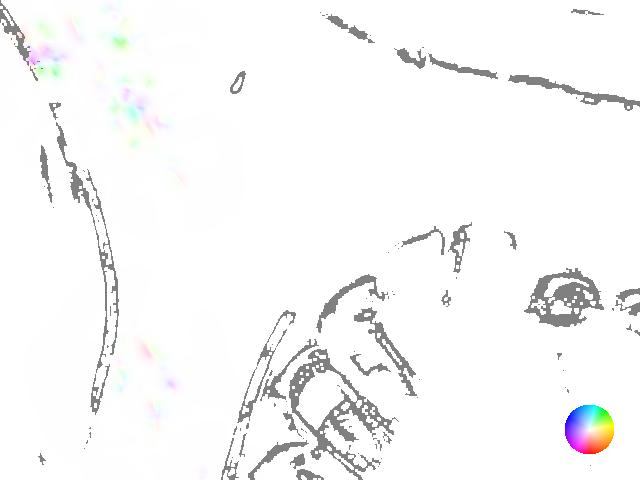} \\ \hline\noalign{\smallskip}

\end{tabular}
\end{table*}

\begin{algorithm}[h]
\caption{Image-based Top View Transformation}
\label{algretarget}
\begin{algorithmic}
\Require Homography $W_{1...n}$, Camera Frames $I_{t, 1...n},I_{t-1, 1...n}$, Background Frames $I_{bg, 1...n}$
\Ensure Top view visualisation of pedestrian motion $I_{Topview}$\\
$I_{Topview}\gets 0$
\For{$i=1$ to $n$}
\State $I_{OF, i}\gets OpticalFlow(I_{t, i},I_{t-1, i})$
\State $I_{Objects, i}\gets ImAbsDiff(I_{t, i}-I_{BG, i})$
\State $I_{Edges, i}\gets Edge(I_{t, i},Canny)$
\State $I_{MeanOF,i}\gets MeanOpticalFlow(I_{Objects, i},I_{OF, i})$
\State $I_{Areas,i}\gets AngularThreshold(I_{Edges, i},I_{OF, i},I_{Objects, i})$
\State $I_{Topview}\gets I_{Topview} + W_{camToWorld,i}(I_{RemainingAreas, i})$
\EndFor
\State $I_{Topview}\gets 1 .* (I_{Topview} > n-1)$
\end{algorithmic}
\end{algorithm}
\begin{algorithm}[h]
\caption{Mean Optical Flow}
\label{algof}
\begin{algorithmic}
\Require Connected Component Image $I_{Objects}$, Optical Flow Image $I_{OF}$
\Ensure Assignment of a mean flow vector for each connected component $I_{MeanOF}$\\
$I_{MeanOF}\gets 0$
\For{$i=1$ to number of connected components in $I_{Objects}$}
\State$u_i\gets 0$, $v_i\gets 0$
\State$counter \gets 0$
\For{each pixel $p$ in $I_{Objects}$ and component $i$}
\State$u_i\gets u_i + u(p)$, $v_i\gets v_i + v(p)$
\State$counter \gets counter +1 $
\EndFor
\State$u_i\gets u_i / counter$,$v_i\gets v_i / counter$
\For{each pixel $p$ in $I_{MeanOF}$ and component $i$}
\State[Assuming components of $I_{Objects}$]
\State$u(p)\gets u_i$, $v(p)\gets v_i$
\State$counter \gets counter +1 $
\EndFor
\EndFor
\end{algorithmic}
\end{algorithm}

\begin{algorithm}[h]
\caption{AngularThreshold}
\label{algat}
\begin{algorithmic}
\Require Edge Image $I_{Edges}$, Connected Component Image $I_{Objects}$, Optical Flow Image $I_{OF}$ 
\Ensure Assignment of remaining areas to be mapped to top view for each connected component $I_{Areas}$\\
$I_{Areas}\gets 0$
\For{$i=1$ to number of connected components in $I_{Objects}$}
\For{each edge pixel $p$ in $I_{Edges}$ and component $i$}
\State $dx(p) \gets \frac{d}{dx}I_{Edges}$ at $p$
\State $dy(p) \gets \frac{d}{dy}I_{Edges}$ at $p$
\If{ (atan2($dy(p)-v(p)$,$dx-u(p)$)$<$ threshold)}
\State $I_{Areas}\gets 1$
\EndIf
\EndFor
\For{each pixel $p$ in $I_{Areas}$ and component $i$}
\State FloodFill in Line-Sweep Manner if there are two pixels $p',p''$ with value 1 in the same horizontal line in $I_{Areas}$
\EndFor
\EndFor
\end{algorithmic}
\end{algorithm}

\begin{figure}
\centering
\includegraphics[height=0.13\textwidth]{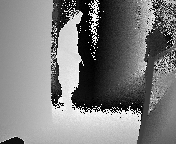}
  \includegraphics[height=0.13\textwidth]{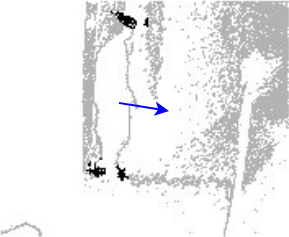}
\includegraphics[height=0.13\textwidth]{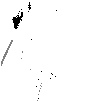}
\caption{A pedestrian captured by a Time-Of-Flight camera (first) is mapped to a top-view by finding edge segments colinear with the mean optical flow vector of the pedestrian (second) to get the overlay in the top view (third).}
\label{fig:4}       % Give a unique label
\end{figure}

In order to make the algorithm applicable for depth-sensor data as well a few alterations are necessary. First, the edge detection is performed
on a single channel image and the contrast is at contact borders (e.g., when a foot is rested on the ground), Fig.~\ref{fig:4}.. Second, the reprojection $W_v: v \in V_{in, depth} \rightarrow v_{\text{topview}} \in V_{\text{target}}$ to the topview
can be implemented by exploiting the camera calibration and the depth values, which avoids distortion errors as described in the previous case.
Thus, Alg. ~\ref{algdepthretarget} is implemented to search for edges in the depth maps and then proceed as described in Alg. \ref{algretarget}. In the final step it uses the camera extrinsics and the depth values of interest, to project them onto $xz$-plane, the ground plane. This way, the intersection of pixel sets from several depth sensors with overlapping viewing cones is not necessary in the final step.
\begin{algorithm}[h]
\caption{Depth-based Top View Transformation}
\label{algdepthretarget}
\begin{algorithmic}
\Require Camera Calibration matrix $C_{depthCam, 1...n}$, Camera Frames $I_{t, 1...n},I_{t-1, 1...n}$, Background Frames $I_{bg, 1...n}$
\Ensure Top view visualisation of pedestrian motion $I_{Topview}$\\
$I_{Topview}\gets 0$
\For{$i=1$ to $n$}
\State $I_{OF, i}\gets OpticalFlow(I_{t, i},I_{t-1, i})$
\State $I_{Objects, i}\gets ImAbsDiff(I_{t, i}-I_{BG, i})$
\State $I_{Edges, i}\gets Edge(I_{t, i},Canny)$
\State $I_{MeanOF,i}\gets MeanOpticalFlow(I_{Objects, i},I_{OF, i})$
\State $I_{Areas,i}\gets AngularThreshold(I_{Edges, i},I_{OF, i},I_{Objects, i})$
\State [Reproject Depth Values]
\State $I_{Topview}\gets I_{Topview} + C_{camToWorld,i}(I_{RemainingAreas, i}).* (1,0,1)$
\State [Discard Height Value]
\EndFor
\State $I_{Topview}\gets 1 .* (I_{Topview} > n-1)$
\end{algorithmic}
\end{algorithm}

\begin{figure}
\centering
\includegraphics[height=0.13\textwidth]{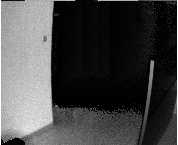}
  \includegraphics[height=0.13\textwidth]{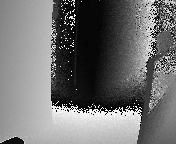}
  \includegraphics[height=0.13\textwidth]{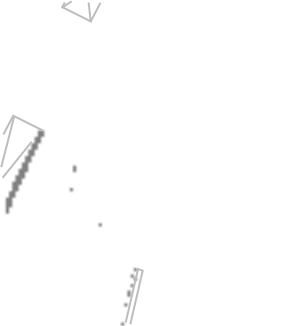}

\caption{Input footage from a calibrated Time-Of-Flight Sensor (intensity, left, depth value, middle) and a topview of the indoor scene cpatured by it (right).}
\label{fig:3}       % Give a unique label
\end{figure}

The altered algorithm is applied to the indoor sequence~\cite{li2012pedestrian} captured with depth sensors as depicted in Table~\ref{tab:2}. The dataset both contains depth images and graylevel images. For our algorithm only used the depth images. The dataset was recorded with the Time-Of-Flight camera SwissRanger SR4000 from Mesa Imaging AG. Beside depth images, the camera can also capture synchronized intensity images. The resolutions of the depth images and intensity images are all 176 x 144 pixels and the distance was ranging from 0 to 5 meters. Using the TOF camera, 4637 pedestrian images and 198 non-pedestrian images in 3 different indoor environments have been captured. Pedestrians in these images are all standing or walking. The body orientation to the camera has not been limited and appear arbitrary in the sequences. A groundtruth  dataset has been made available containing manually labeled pedestrian positions.  The 4637 pedestrian (positive) samples
have originally been divided into two parts for training and testing purposes in pedestrian recognition:
There are 3160 positive training samples, and 1477 positive test samples.  Again, we apply the motion analysis by computing the cumulative grid, Table \ref{tab:2}  third row, and the optical flow, Table \ref{tab:2}  fourth row.
\begin{table*}[t] 
\caption{Analysis and summarization of an indoor pedestrian database~\cite{li2012pedestrian}. The input data are captured with a Time-Of-Flight Camera.}
\centering
\label{tab:2}       % Give a unique label
\begin{tabular}{cccccccc}
\hline\noalign{\smallskip}
\includegraphics[width=0.1\linewidth]{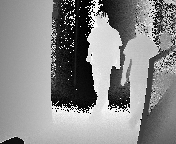} & \includegraphics[width=0.1\linewidth]{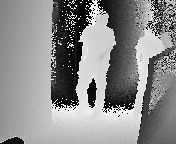} &\includegraphics[width=0.1\linewidth]{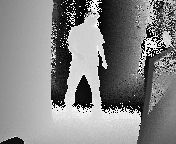} &\includegraphics[width=0.1\linewidth]{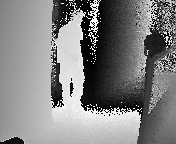} &\includegraphics[width=0.1\linewidth]{depth11.png}&\includegraphics[width=0.1\linewidth]{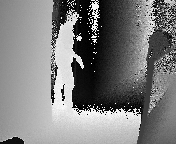}&\includegraphics[width=0.1\linewidth]{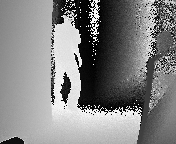}&\includegraphics[width=0.1\linewidth]{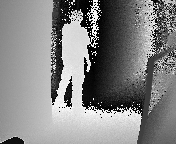}\\ \hline\noalign{\smallskip}
$t$&$t+1$&$t+2$&$t+3$&$t+4$&$t+5$&$t+6$&$t+7$\\
\\ \hline\noalign{\smallskip}
\\
\includegraphics[width=0.1\linewidth]{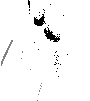} & \includegraphics[width=0.1\linewidth]{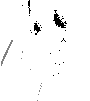} &\includegraphics[width=0.1\linewidth]{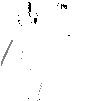} &\includegraphics[width=0.1\linewidth]{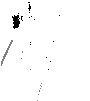} &\includegraphics[width=0.1\linewidth]{topview11.png}&\includegraphics[width=0.1\linewidth]{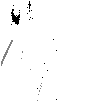}&\includegraphics[width=0.1\linewidth]{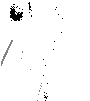}&\includegraphics[width=0.1\linewidth]{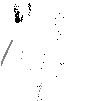}\\
\includegraphics[width=0.1\linewidth]{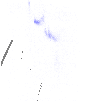} & \includegraphics[width=0.1\linewidth]{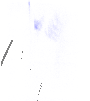} &\includegraphics[width=0.1\linewidth]{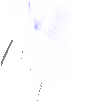} &\includegraphics[width=0.1\linewidth]{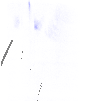} &\includegraphics[width=0.1\linewidth]{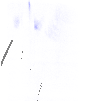}&\includegraphics[width=0.1\linewidth]{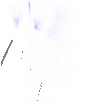}&\includegraphics[width=0.1\linewidth]{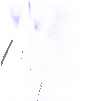}&\includegraphics[width=0.1\linewidth]{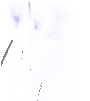}\\ 
\includegraphics[width=0.1\linewidth]{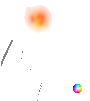} & \includegraphics[width=0.1\linewidth]{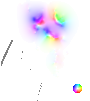} &\includegraphics[width=0.1\linewidth]{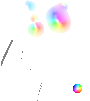} &\includegraphics[width=0.1\linewidth]{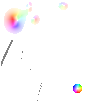} &\includegraphics[width=0.1\linewidth]{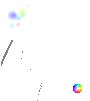}&\includegraphics[width=0.1\linewidth]{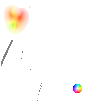}&\includegraphics[width=0.1\linewidth]{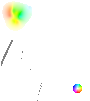}&\includegraphics[width=0.1\linewidth]{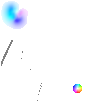}\\ 
\\ \hline\noalign{\smallskip}
\end{tabular}
\end{table*}

In a third example, RGB-Cameras are used in an indoor surveillance scenario, the capturing of pedestrian motion in a train station~\cite{pets07}. The calibration of the cameras has been conduted with Tsais Algorithm by relying on equidistant markers that are placed on the floor of the terminal. The distance between the markers was 1.8m (to a tolerance of $\pm1$cm). All spatial measurements have been set in metres, again. For recording, the following cameras were used: 2x Canon MV-1 1xCCD w/progressive scan, 2x Sony DCR-PC1000E 3xCMOS (full colour, 768 x 576 pixels, 25 frames per second). The video footage has been compressed as JPEG image sequences (approx. 90\% quality). The dataset displays loitering behaviour (a person standing still for longer than 60 seconds), theft of belongings and leaving luggage behind. While the lighting conditions remained stable, the density of the crowd varied with each sequence.
Its outcome is visualized in Table~\ref{tab:3}. 
\begin{table*}[t]
\caption{Analysis and summarization of an indoor pedestrian database~\cite{pets07}. The crowding that is noticable at time frames $t+2,t+3,t+4$ in the input images (first row) can be directly read from the cumulative top view scatterplot (third row). An optical flow computation can be applied to the top view, e.g. to discriminate pixel movements or search for unusual motion patterns.}
\centering
\label{tab:3}       % Give a unique label
\begin{tabular}{cccccccc}
\hline\noalign{\smallskip}
\includegraphics[width=0.1\linewidth]{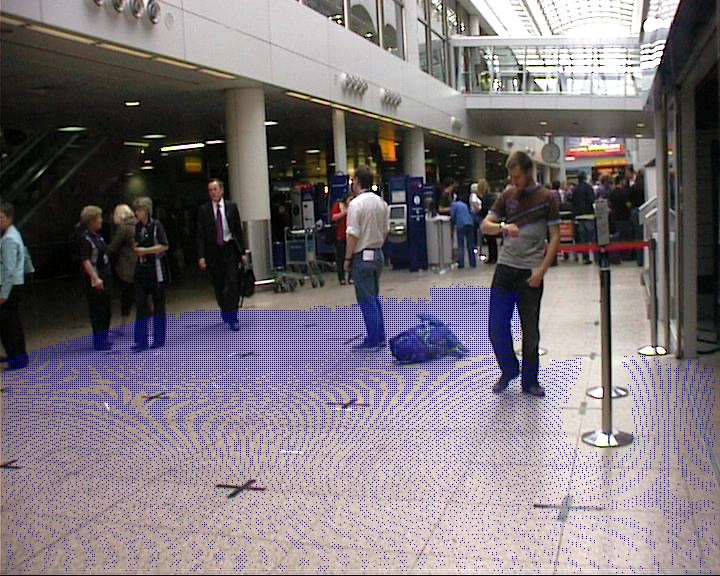} & \includegraphics[width=0.1\linewidth]{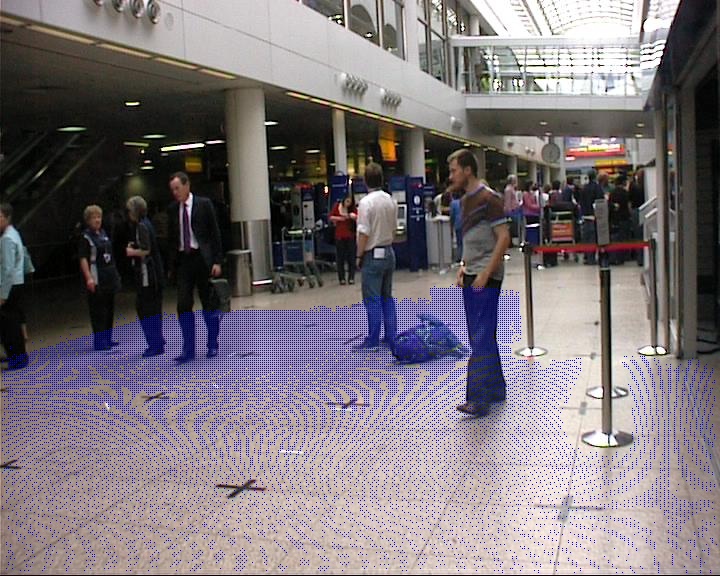} &\includegraphics[width=0.1\linewidth]{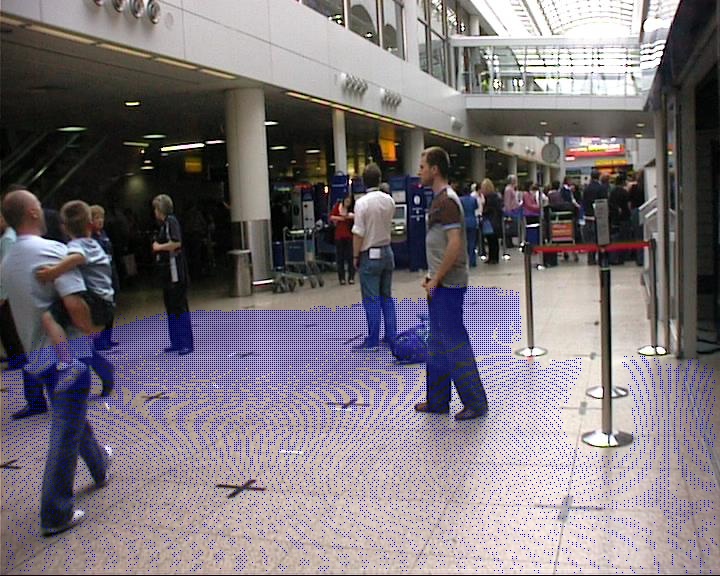} &\includegraphics[width=0.1\linewidth]{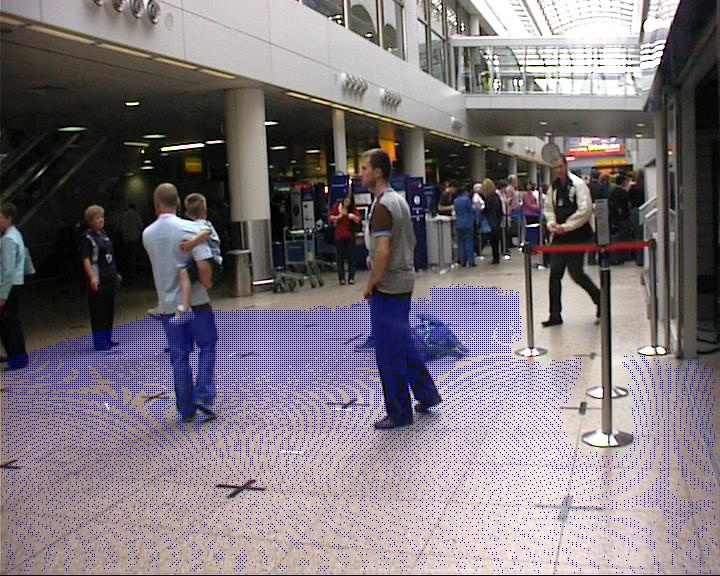} &\includegraphics[width=0.1\linewidth]{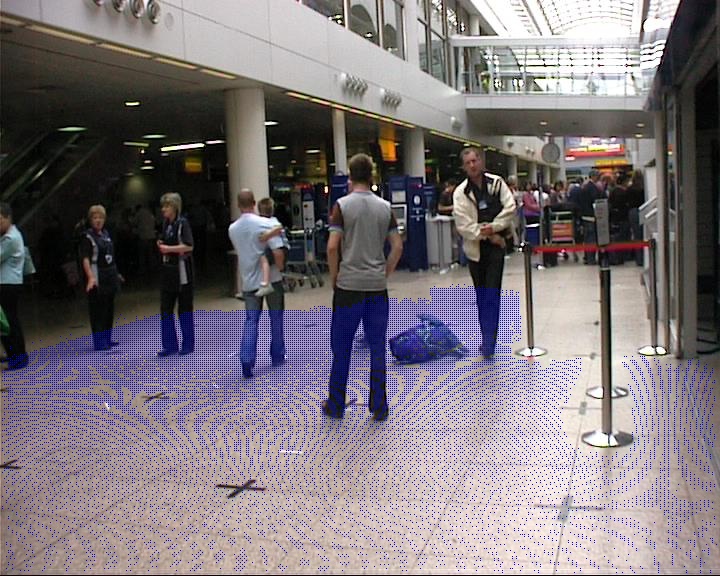}&\includegraphics[width=0.1\linewidth]{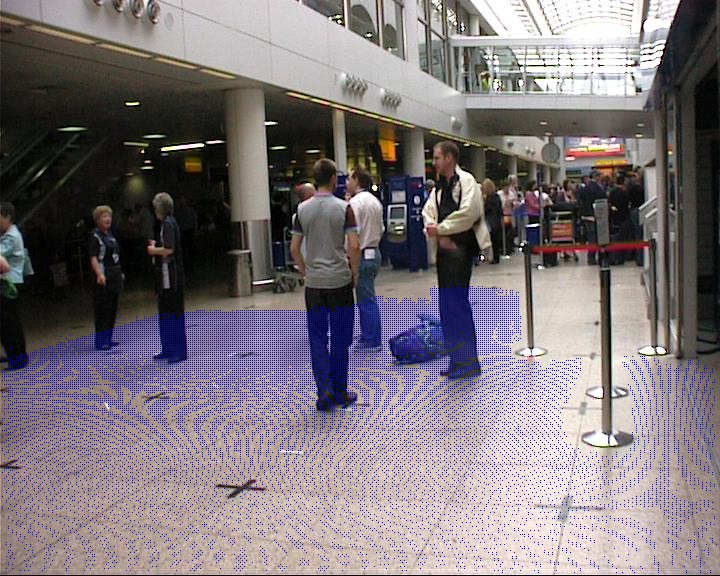}&\includegraphics[width=0.1\linewidth]{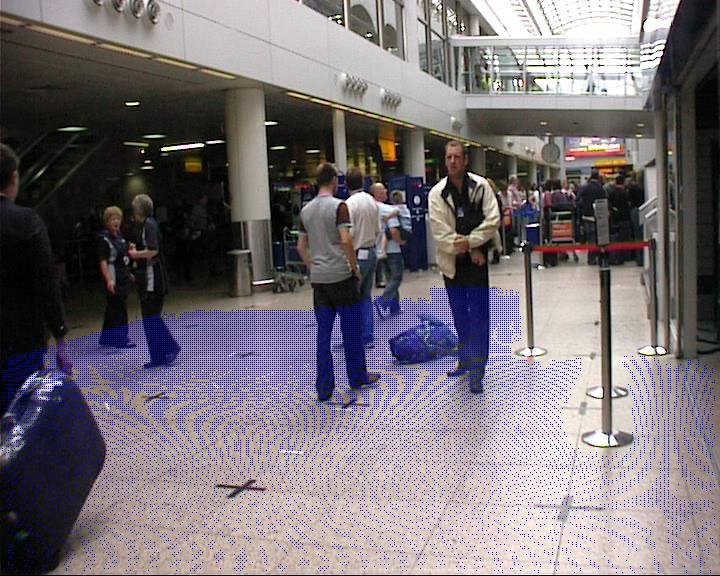}&\includegraphics[width=0.1\linewidth]{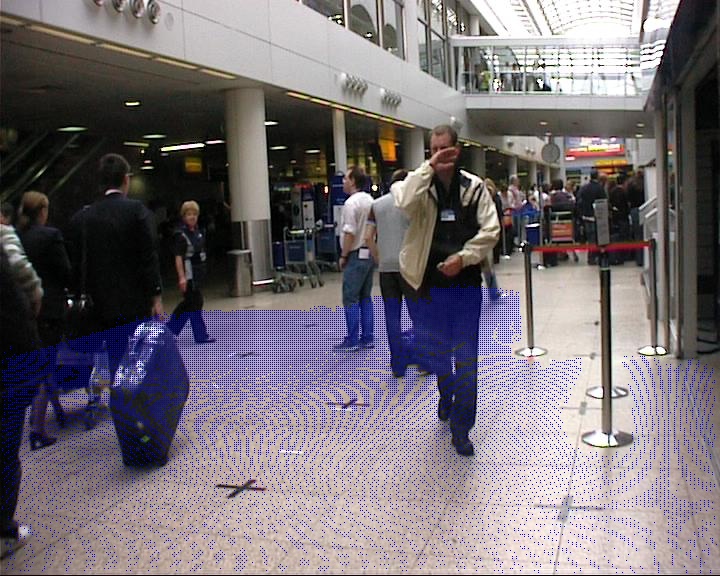}\\ \hline\noalign{\smallskip}
$t$&$t+1$&$t+2$&$t+3$&$t+4$&$t+5$&$t+6$&$t+7$\\
\\ \hline\noalign{\smallskip}
\\
\includegraphics[width=0.1\linewidth]{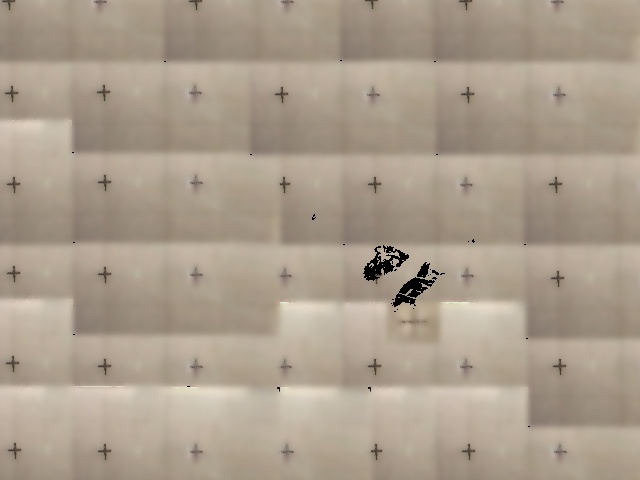} & \includegraphics[width=0.1\linewidth]{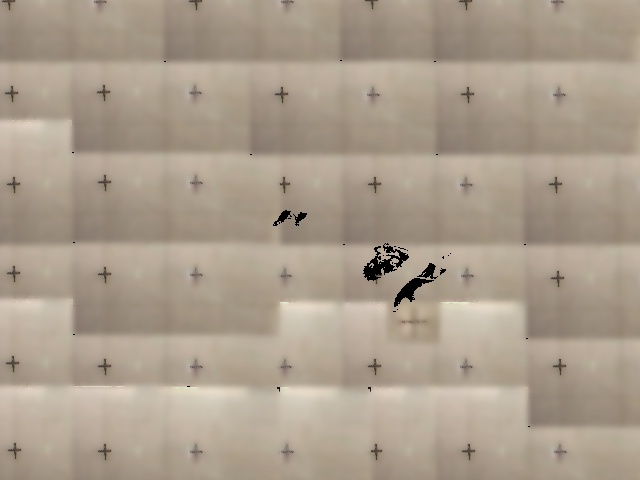} &\includegraphics[width=0.1\linewidth]{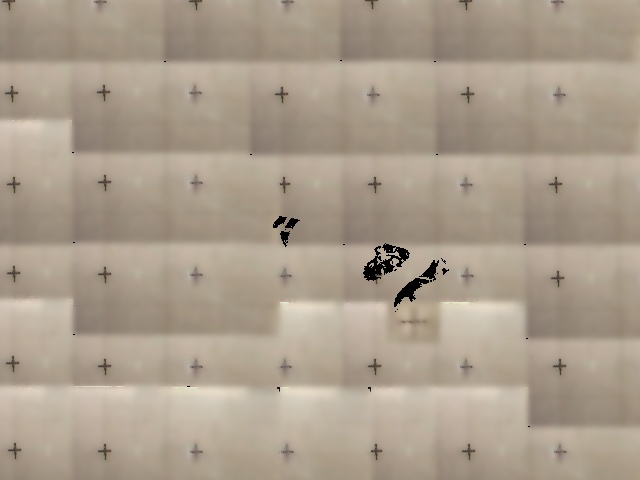} &\includegraphics[width=0.1\linewidth]{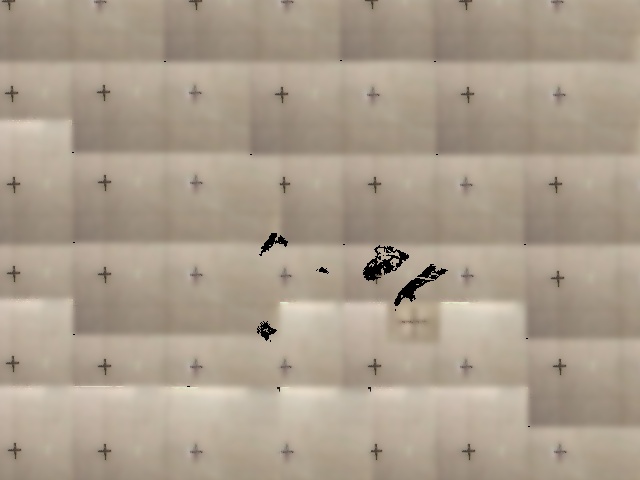} &\includegraphics[width=0.1\linewidth]{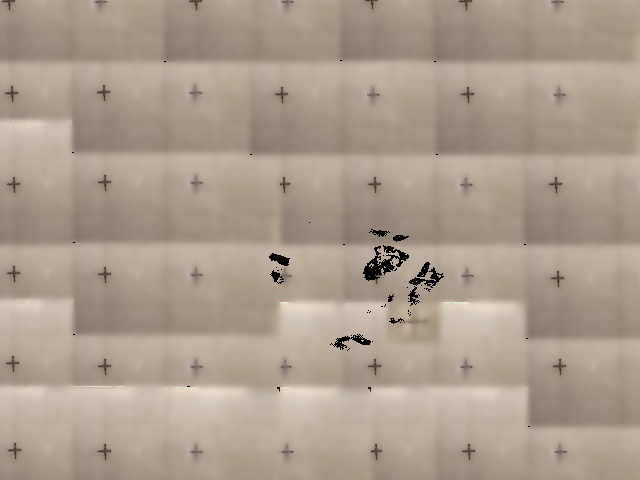}&\includegraphics[width=0.1\linewidth]{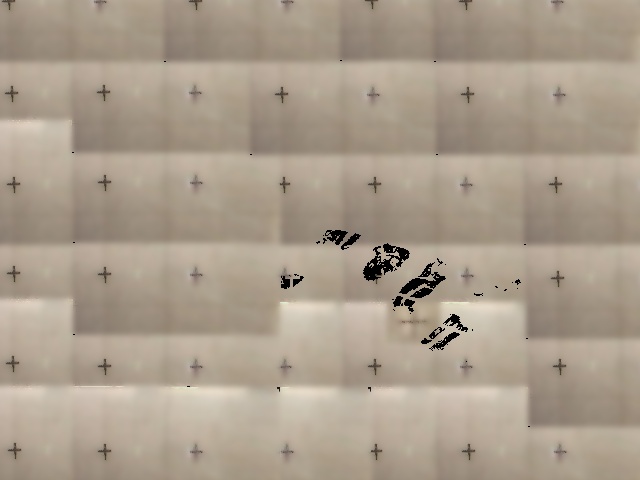}&\includegraphics[width=0.1\linewidth]{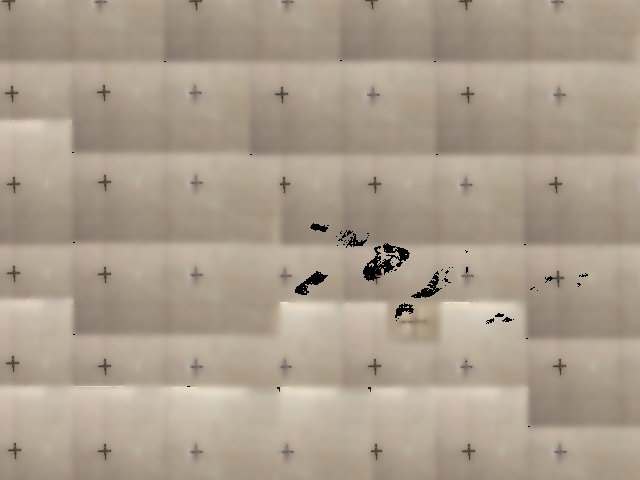}&\includegraphics[width=0.1\linewidth]{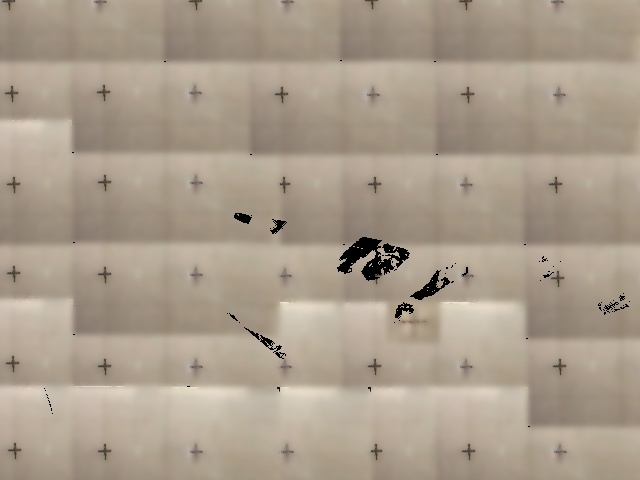} \\
\includegraphics[width=0.1\linewidth]{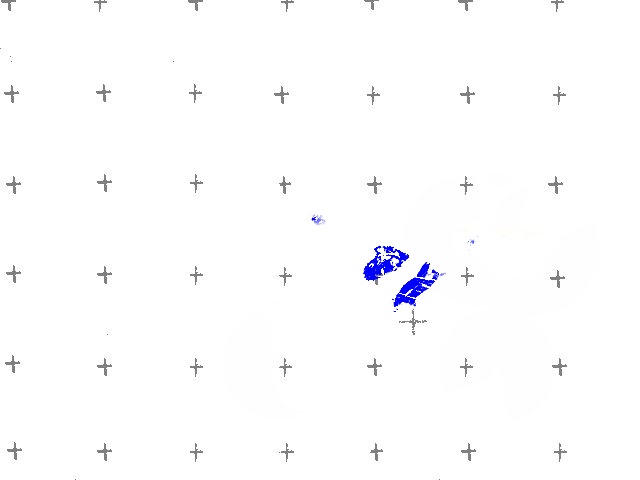} & \includegraphics[width=0.1\linewidth]{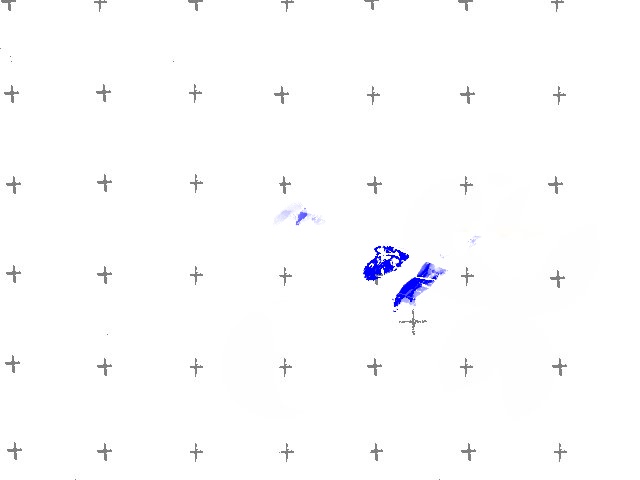} &\includegraphics[width=0.1\linewidth]{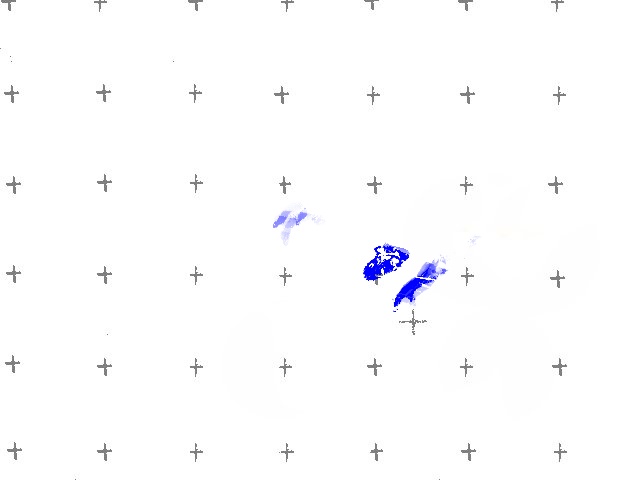} &\includegraphics[width=0.1\linewidth]{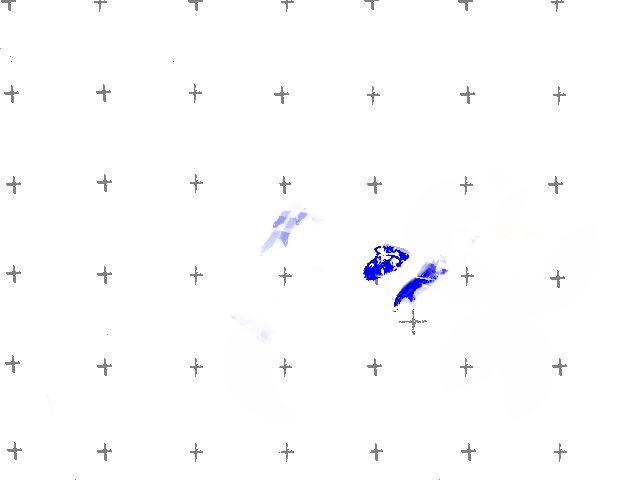} &\includegraphics[width=0.1\linewidth]{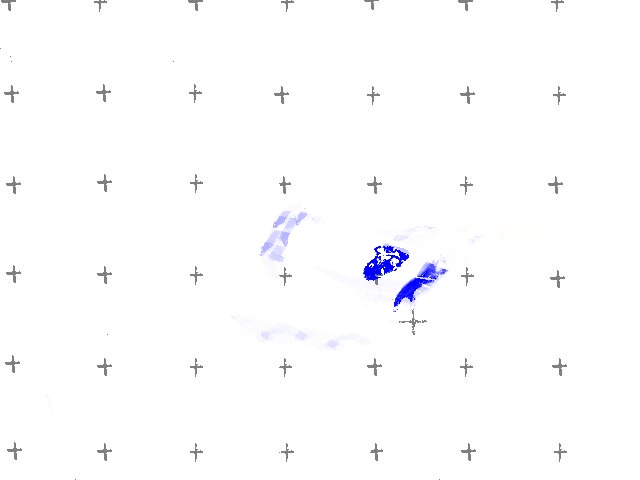}&\includegraphics[width=0.1\linewidth]{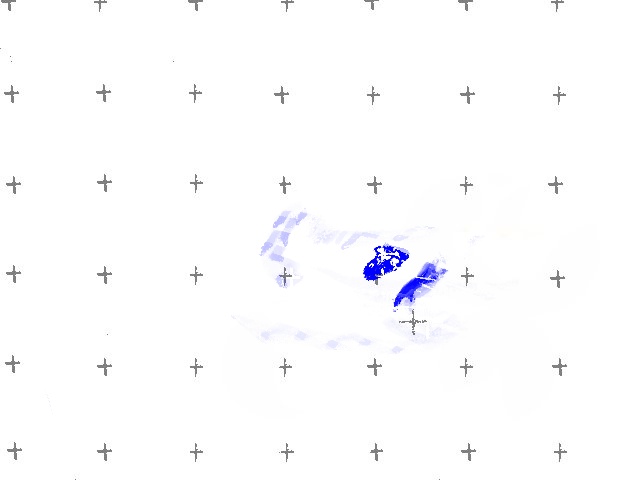}&\includegraphics[width=0.1\linewidth]{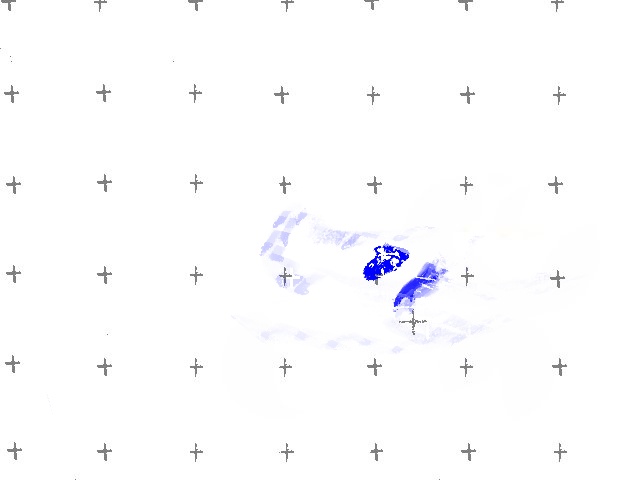}&\includegraphics[width=0.1\linewidth]{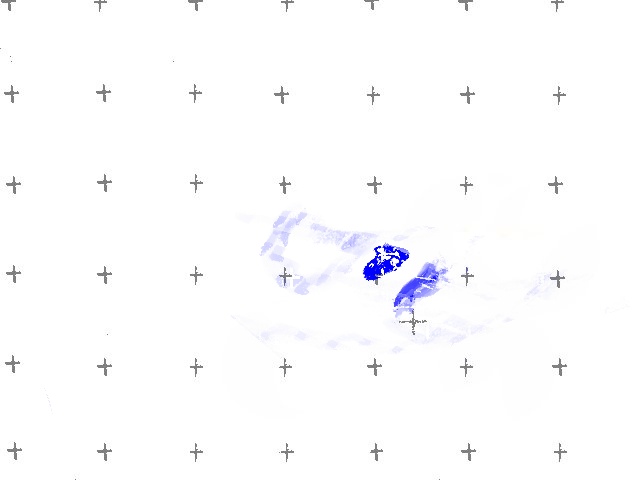} \\\includegraphics[width=0.1\linewidth]{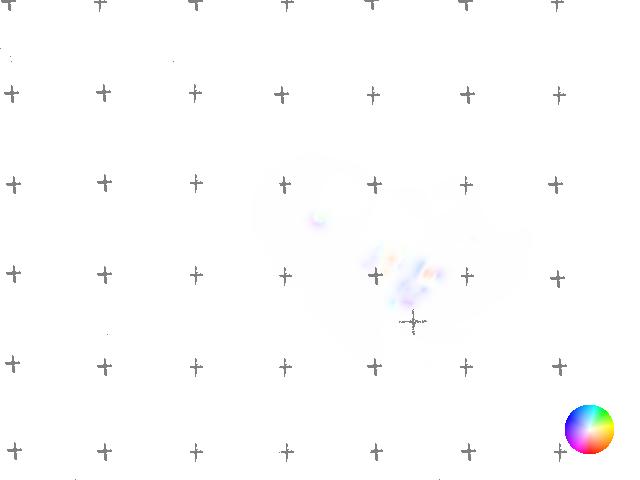} &\includegraphics[width=0.1\linewidth]{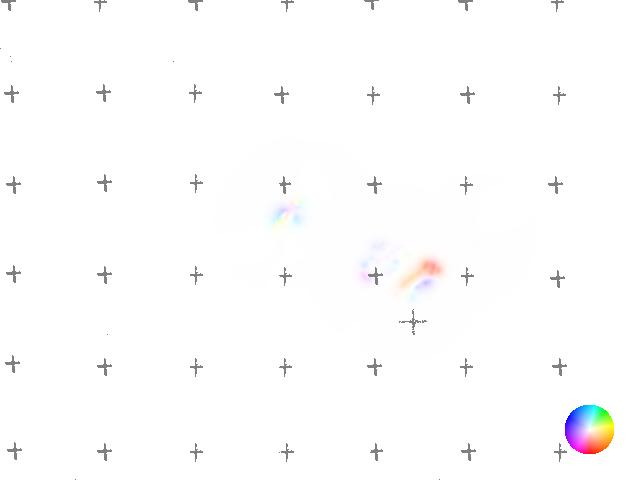} &\includegraphics[width=0.1\linewidth]{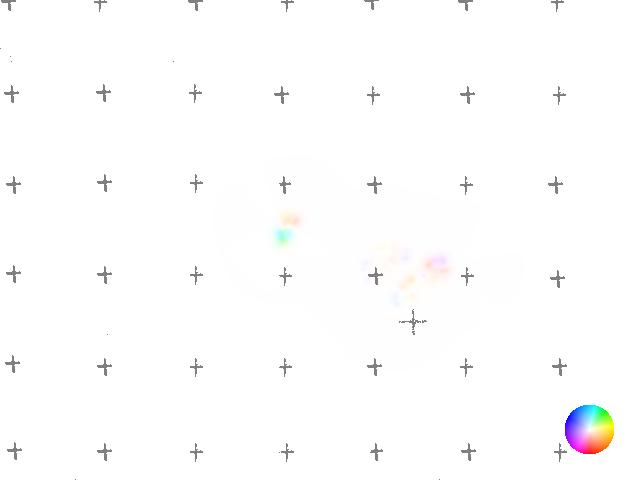}&\includegraphics[width=0.1\linewidth]{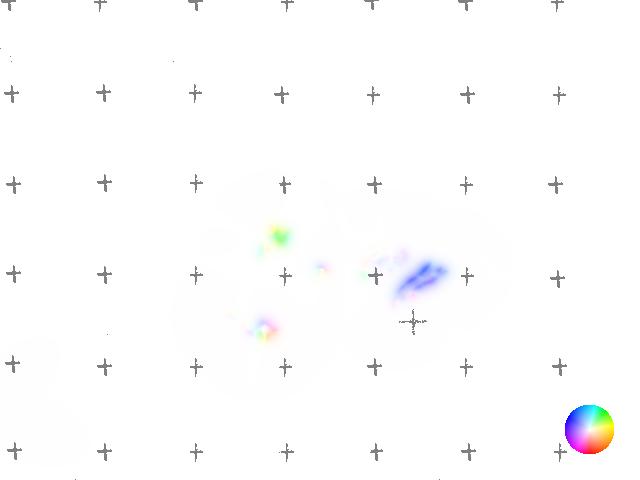} &\includegraphics[width=0.1\linewidth]{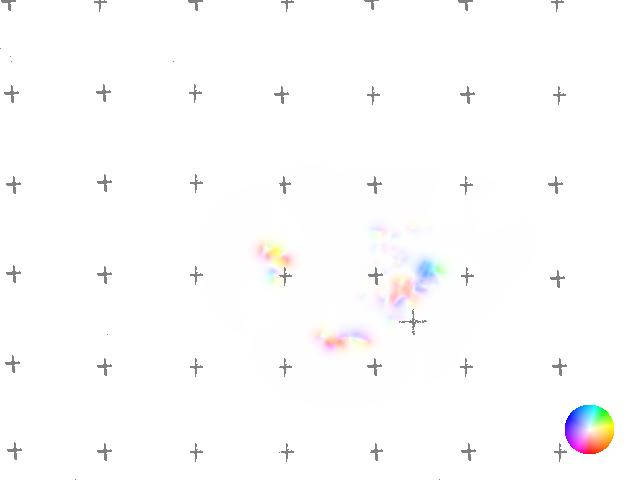}&\includegraphics[width=0.1\linewidth]{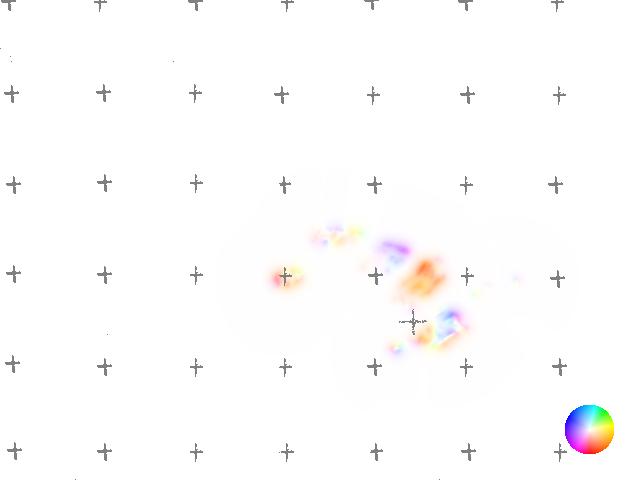}&\includegraphics[width=0.1\linewidth]{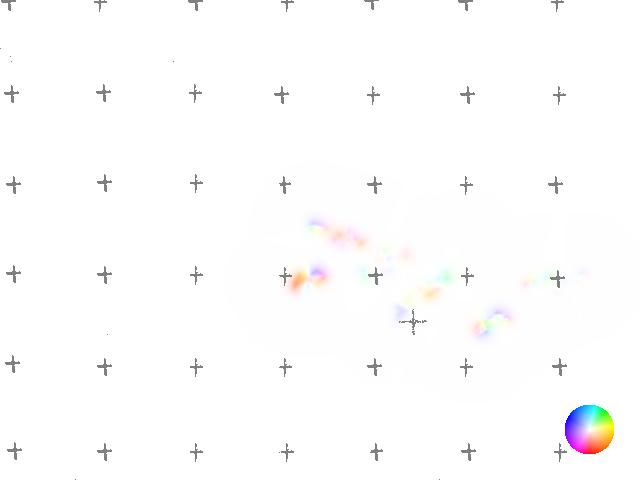}&\includegraphics[width=0.1\linewidth]{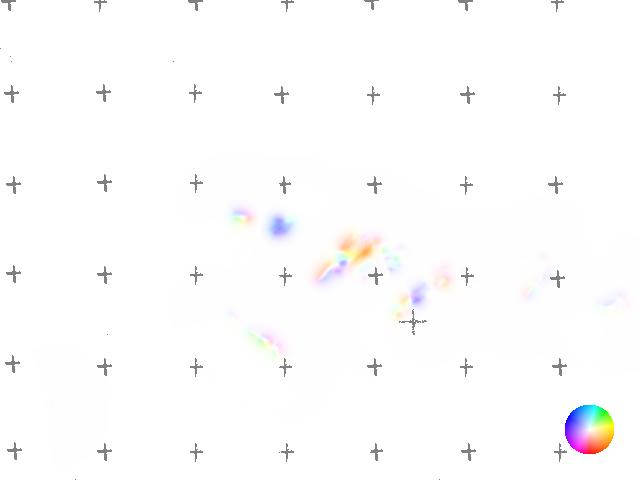} \\ \hline\noalign{\smallskip}

\end{tabular}
\end{table*}
\section{Results}
\label{results}
We have applied the algorithm described in the previous section to all three datasets to retrieve cumulative grids and optical flow maps in the topview.
The results for the PETS dataset 2009~\cite{pets09} are listed in Table \ref{tab:resultspets2009}. In the S1 subsets the crowding behaviour is visible in the cumulative grid. In subset S2.L3
the positions of the two people who remain still throughout the sequence are imaged with saturated dots. In subset S3.MF the motion paths of the two people in yellow vests who approach the crowd 
are imaged as less saturated streaks. 

\begin{figure*}
\centering

  \includegraphics[width=0.3\textwidth]{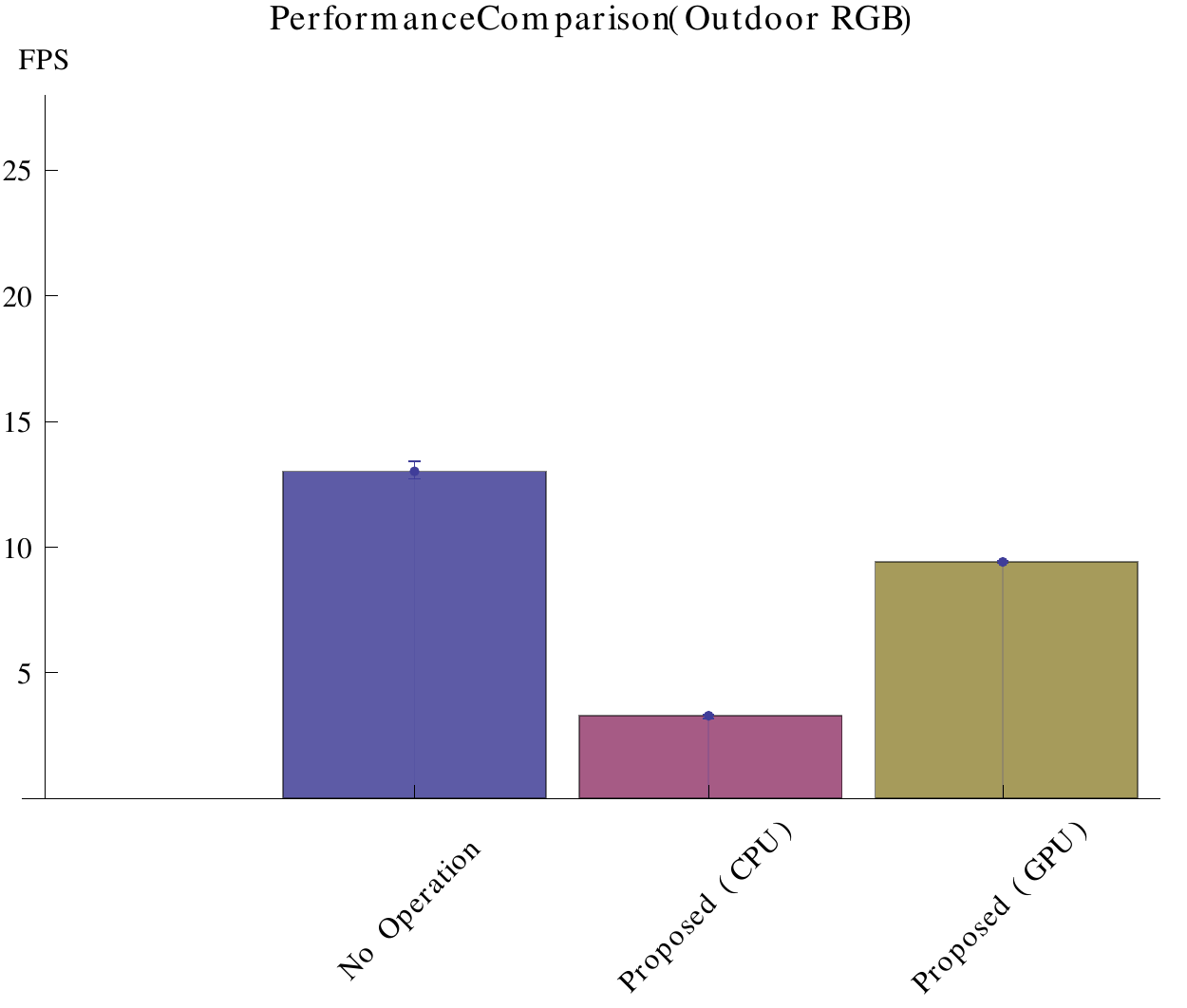}
  \includegraphics[width=0.3\textwidth]{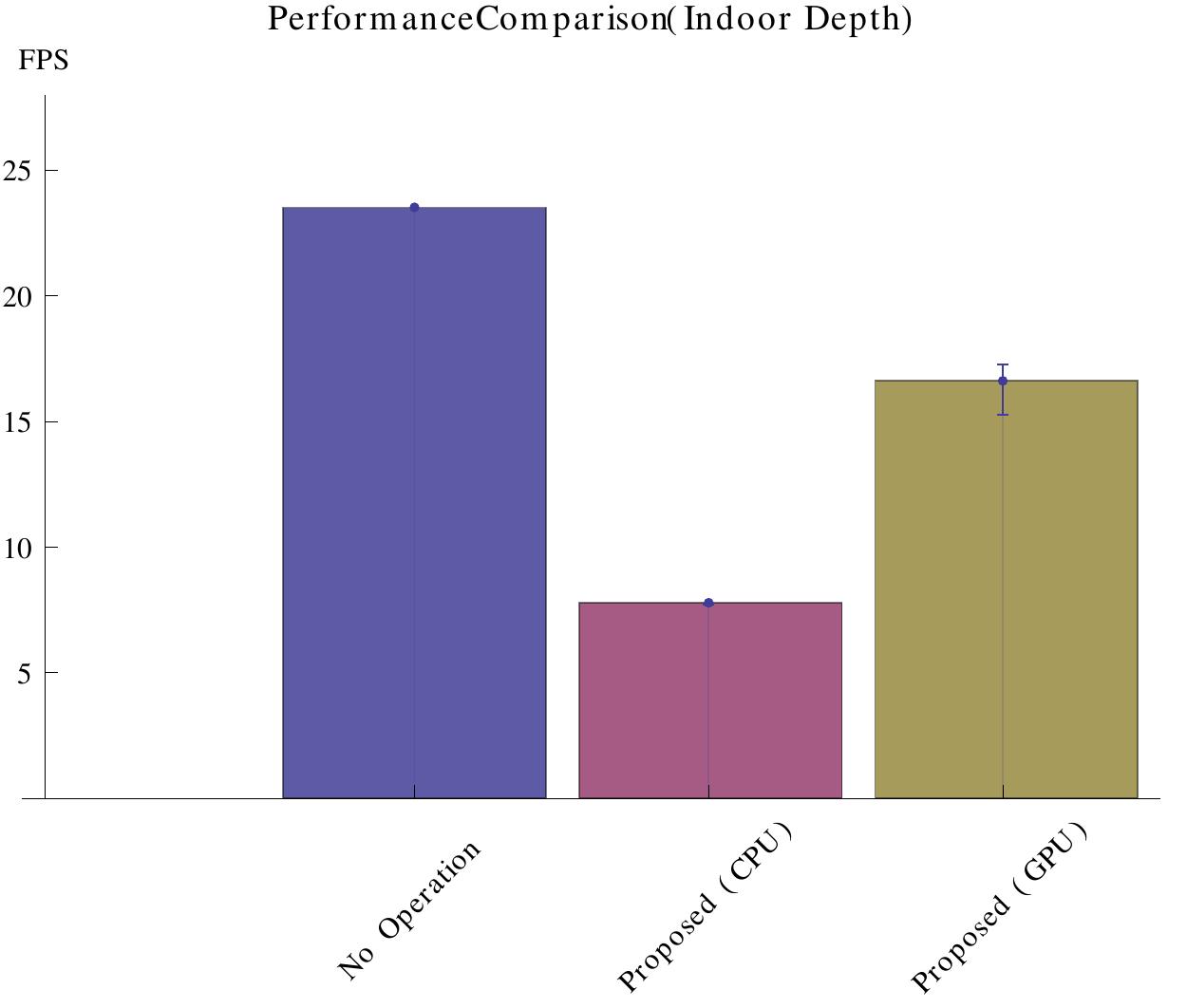}
\includegraphics[width=0.3\textwidth]{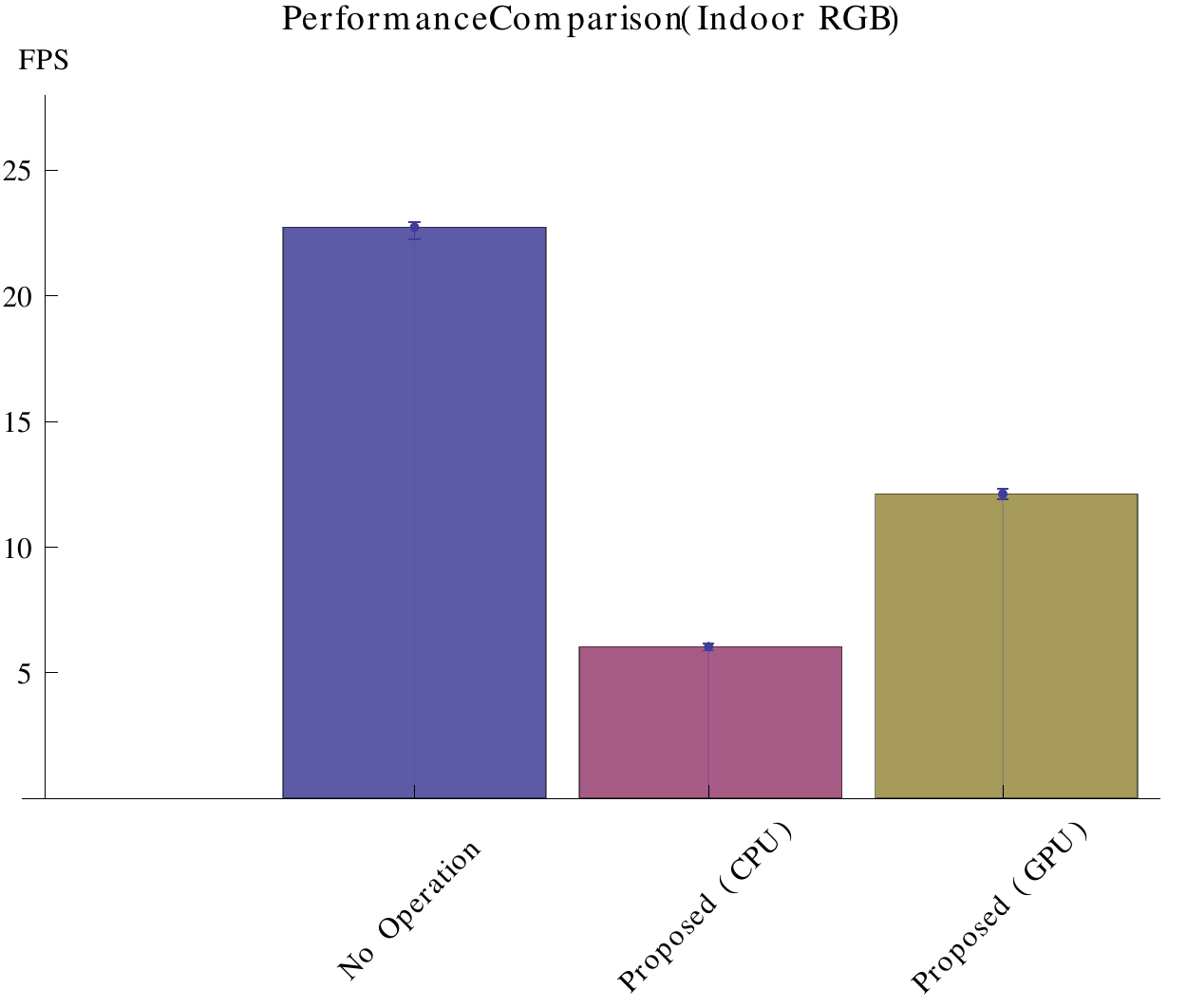}

\caption{The performance analysis for CPU vs. CUDA-accelerated implementations of the algorithm. We tested the algorithm with the three datasets. Since eight cameras have to be processed, the outdoor dataset~\cite{pets09} is the most consuming whereas the indoor depth dataset~\cite{li2012pedestrian} with a single input stream is the least consuming. The GPU based implementation outperforms a purely CPU-based implementation and approximates an idle loop (no processing) more closely.}
\label{fig:perf}       % Give a unique label
\end{figure*}

The results for the Depth Dataset~\cite{li2012pedestrian} are listed in Table \ref{tab:resultsdepth}. In subset S16 the motion patterns of the three people are visible both in the cumulative grid and the optical flow map. The motion pattern of the three people moving far from the camera in subset S9 are well imaged in the optical flow map. The motion patterns of two people crossing each other may confuse the optical flow algorithm, as seen in subset S1.
The results for the PETS dataset 2007~\cite{pets07} are listed in Table \ref{tab:resultspets2007}.
Saturated spots indicate loitering people (S1) or bags which have been dropped (S2-S8). A streak towards the spot indicates suspicious behaviour, such as theft (S5).
We have compared the performance of the algorithm on the datasets for a CPU and GPU implementation, Fig.~\ref{fig:perf}.
The algorithm has been implemented in C++ using the OpenCV library. The CPU code was run on a Intel i7 Dual Core Processor and GPU implementation was run on a
NVIDIA GeForce GTX 690 with 2048Mb Cache, 1536 cores, and NVIDIA Driver 4.20.
We compared the framerate in fps against the routine at idle, i.e. reading in the image streams without processing (\emph{no operation}). It can be seen, that
with the GPU implementation we can achieve real-time framerates (greater than 10 fps) that enable the pedestrian surveillance in top view with life cumulative maps.
As all eight cameras were processed, the outdoor dataset~\cite{pets09} is the most consuming whereas the indoor depth dataset~\cite{li2012pedestrian} with a single input stream is the least consuming.
\def\imagetop#1{\vtop{\null\hbox{#1}}}

\begin{table*}[t]
\caption{The results of our algorithm applied to different sequences of the PETS dataset 2009~\cite{pets09}}
\centering
\label{tab:resultspets2009}       % Give a unique label    
\begin{tabular}{p{1cm}p{1cm}p{3cm}p{2cm}p{2cm}p{2cm}p{3cm}}
\hline
PETS dataset 2009~\cite{pets09}&S1.L1&The sequence contains a medium density crowd, who are running. There are overcast lighting conditions& \imagetop{\includegraphics[height=0.7\linewidth]{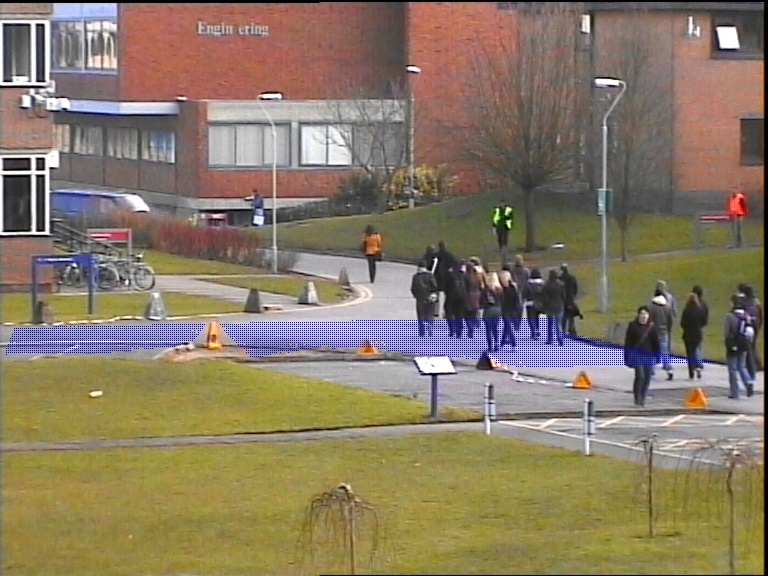}} &\imagetop{\includegraphics[height=0.7\linewidth]{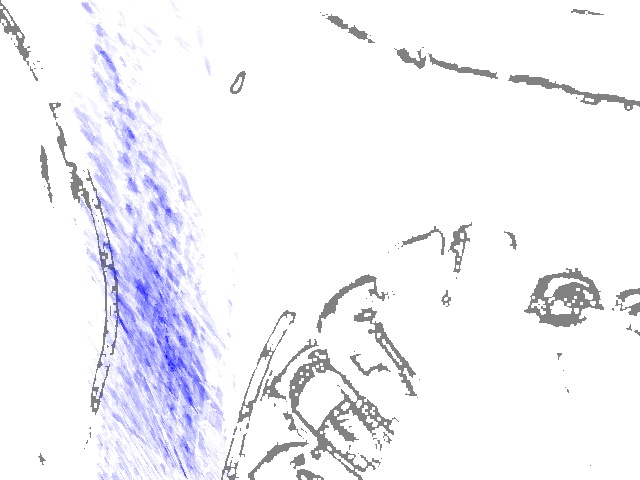}}&\imagetop{\includegraphics[height=0.7\linewidth]{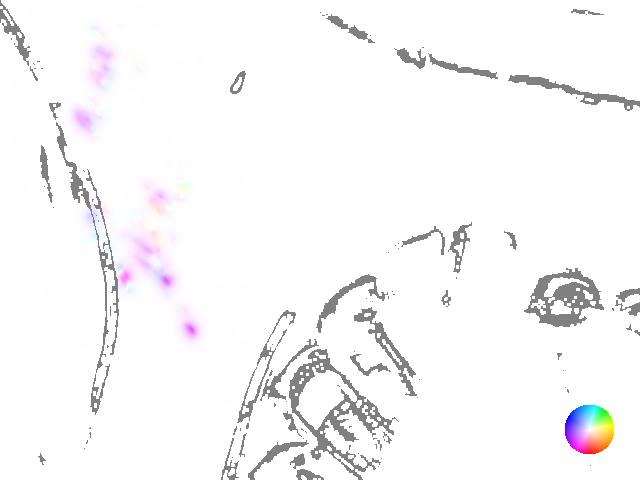}}&The cumulative grid shows the occupied area by the moving crowd in the middle of the sequence.\\
\hline
PETS dataset 2009~\cite{pets09}&S1.L2&The sequence contains a high density crowd, who are walking in diverse directions. There are bright sunshine and shadows	& \imagetop{\includegraphics[height=0.7\linewidth]{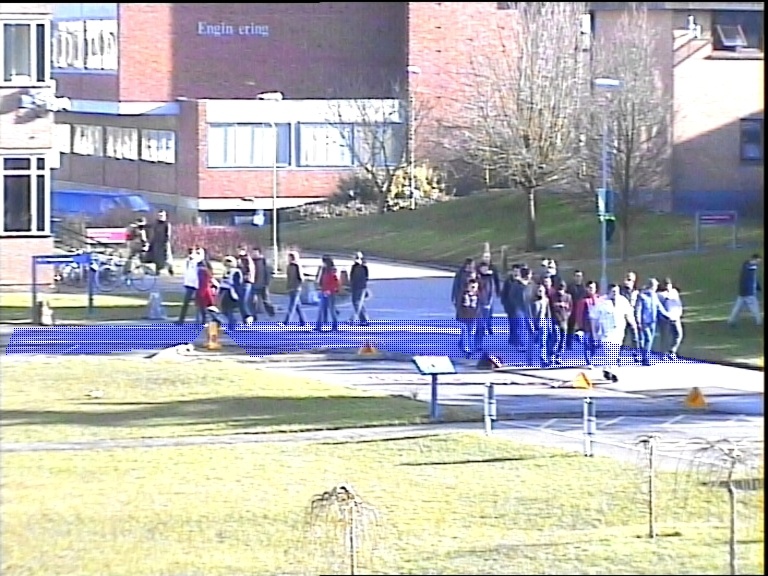}} &\imagetop{\includegraphics[height=0.7\linewidth]{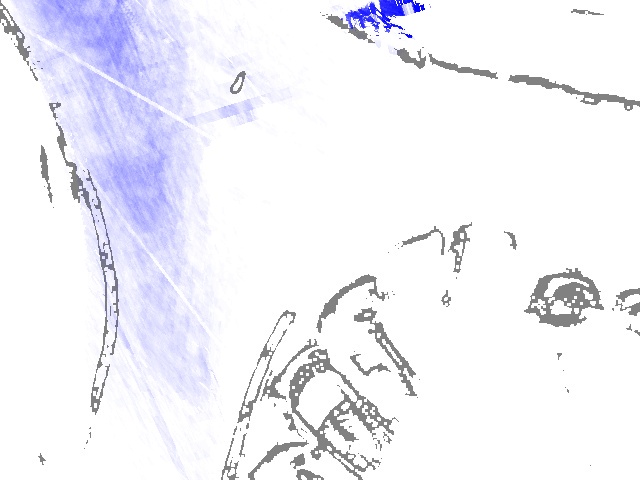}}&\imagetop{\includegraphics[height=0.7\linewidth]{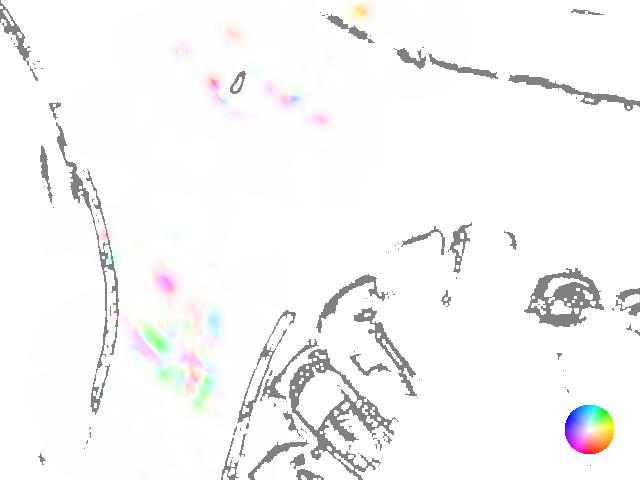}}&The cumulative grid shows two blobs indicating the diversion of the flow.\\
\hline
PETS dataset 2009~\cite{pets09}&S1.L3&The sequence contains a medium density crowd, who are walking. There are bright sunshine and shadows
 visible&\imagetop{\includegraphics[height=0.7\linewidth]{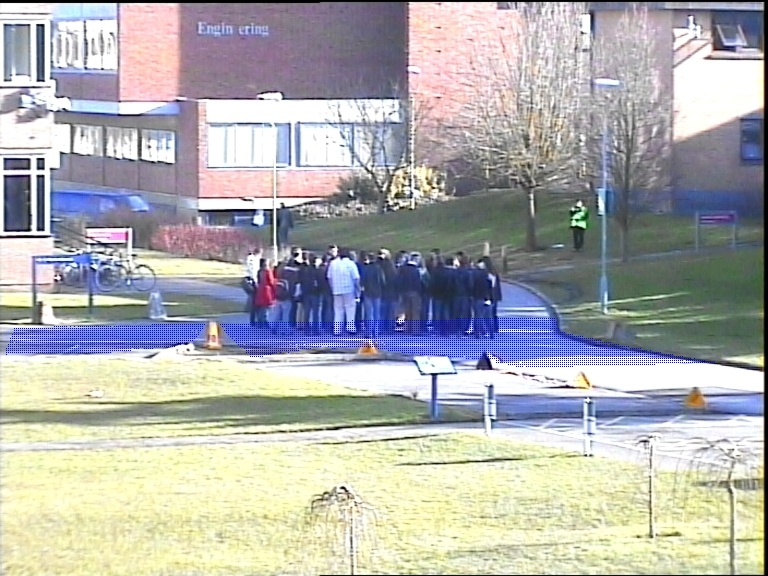}} &\imagetop{\includegraphics[height=0.7\linewidth]{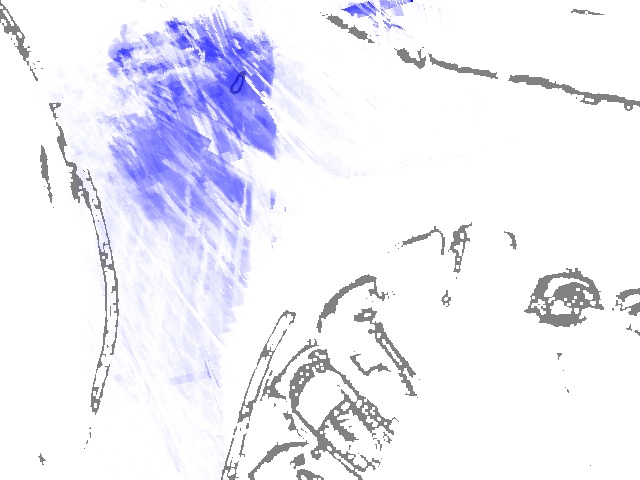}}&\imagetop{\includegraphics[height=0.7\linewidth]{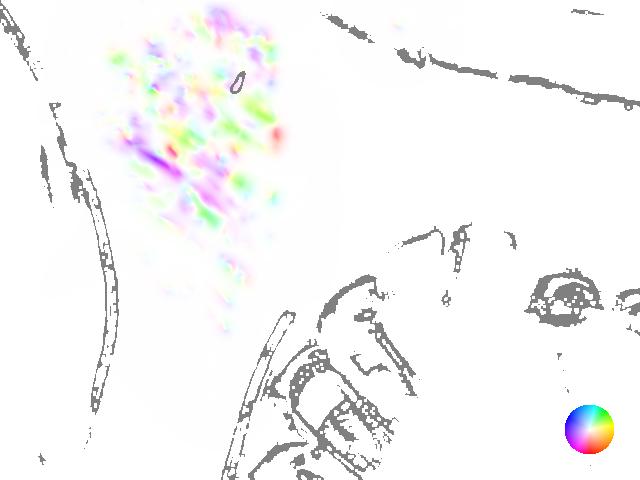}}&The area occupied by the crowd is visible in the cumulative grid in the middle of the sequence.\\
\hline
PETS dataset 2009~\cite{pets09}&S2.L1&The sequence contains a sparse crowd, who are walking. There are overcast lighting conditions
 visible&\imagetop{\includegraphics[height=0.7\linewidth]{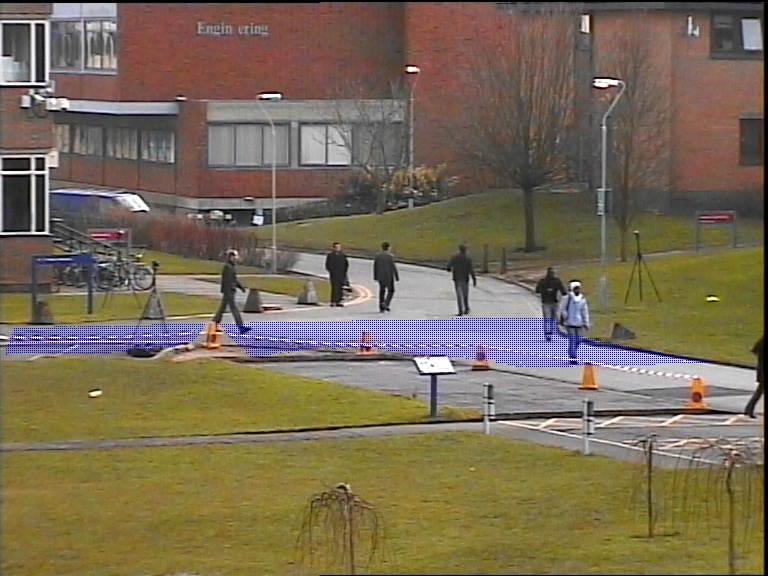}} &\imagetop{\includegraphics[height=0.7\linewidth]{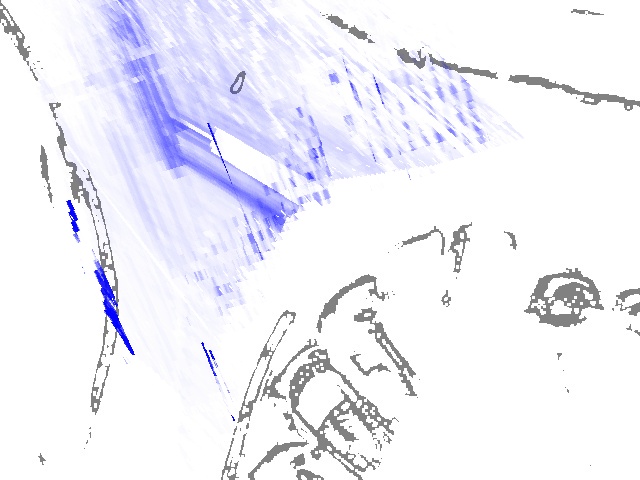}}&\imagetop{\includegraphics[height=0.7\linewidth]{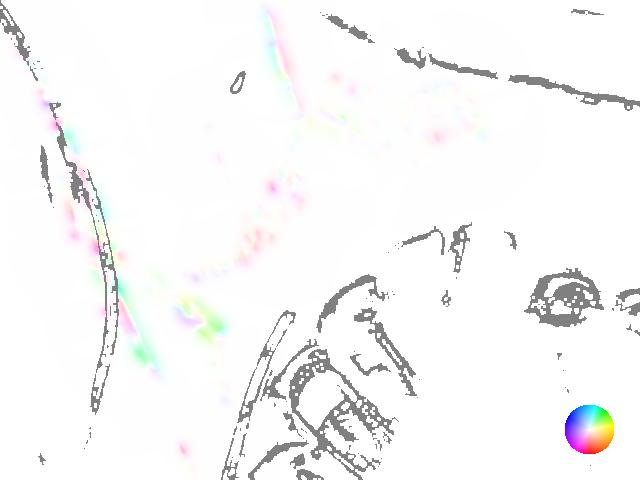}}&The cumulative grid is less dense. Few areas mainly traversed by the pedestrians appear more saturated in the grid.\\
\hline
PETS dataset 2009~\cite{pets09}&S2.L2&The sequence contains a medium density crowd, who are walking. There are bright sunshine and shadows
 visible&\imagetop{\includegraphics[height=0.7\linewidth]{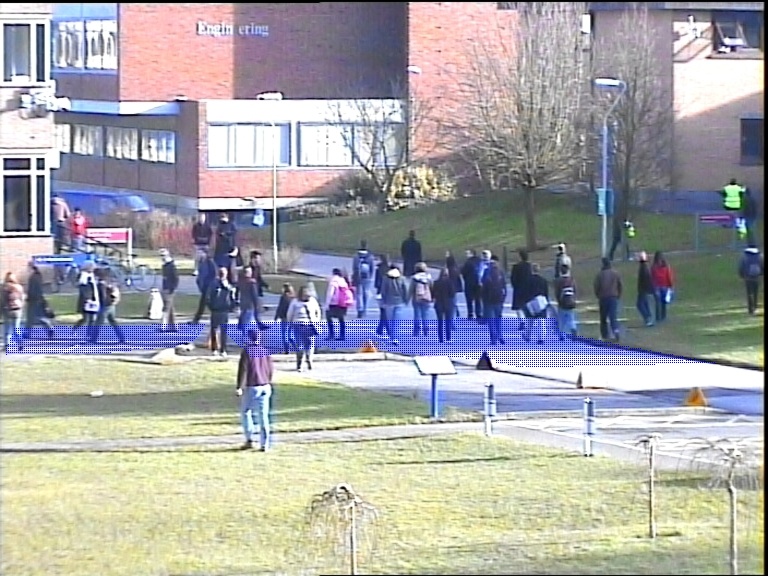}} &\imagetop{\includegraphics[height=0.7\linewidth]{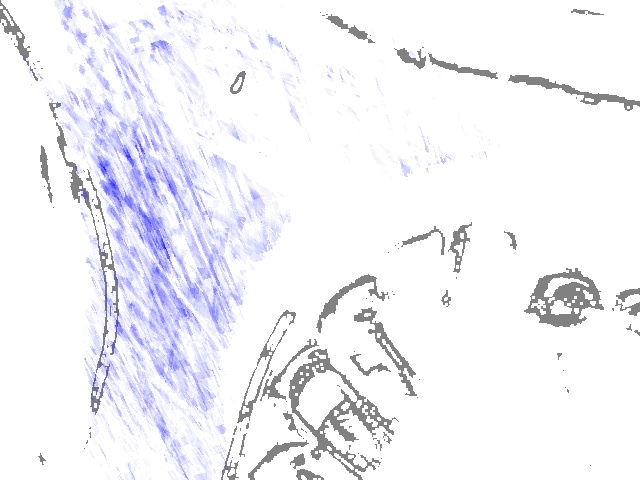}}&\imagetop{\includegraphics[height=0.7\linewidth]{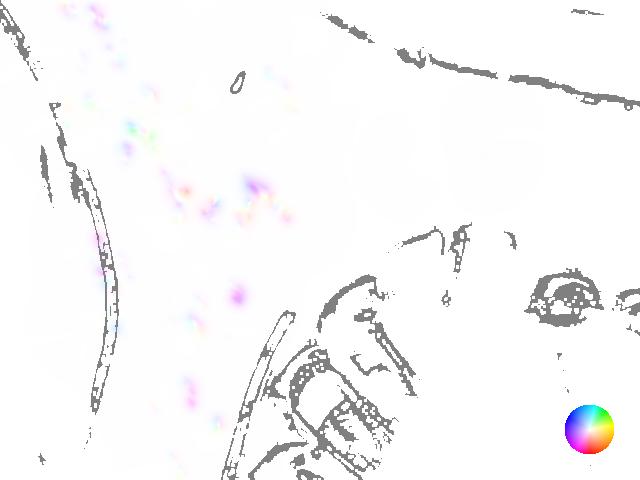}}&The scattered incoherent crowd movements are visible in the less saturated grid.\\
\hline
PETS dataset 2009~\cite{pets09}&S2.L3&The sequence contains a high density crowd, who are walking. There are overcast lighting conditions
 visible&\imagetop{\includegraphics[height=0.7\linewidth]{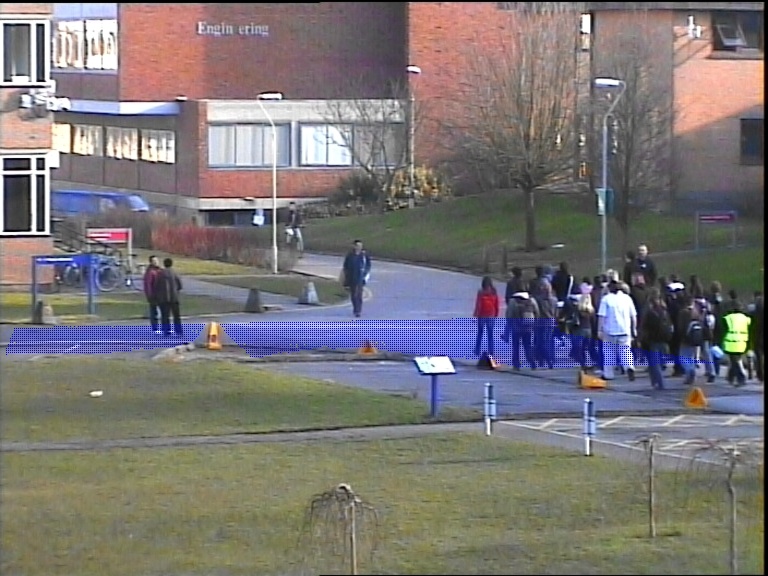}} &\imagetop{\includegraphics[height=0.7\linewidth]{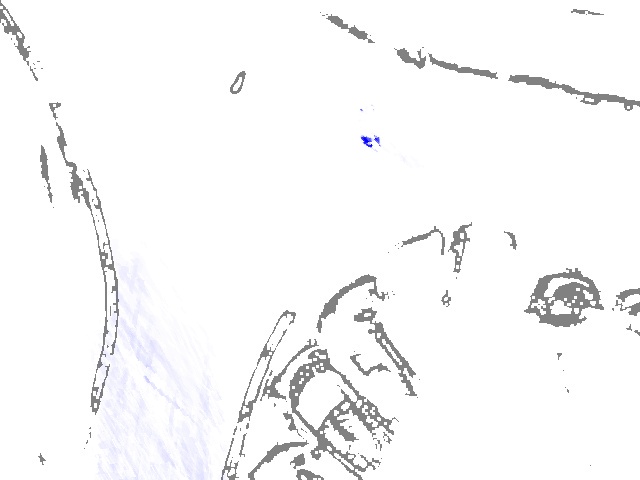}}&\imagetop{\includegraphics[height=0.7\linewidth]{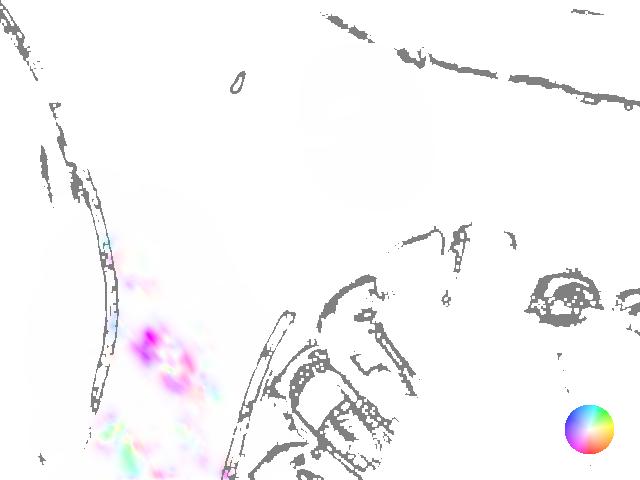}}&In the grid, two saturated spots denote the position of the two men remaining still while the walking crowd is imaged in a less saturated region.\\
\hline
PETS dataset 2009~\cite{pets09}&S3.HL
&The sequence contains a high density crowd, who are running
. There are bright sunshine and shadows
 visible&\imagetop{\includegraphics[height=0.7\linewidth]{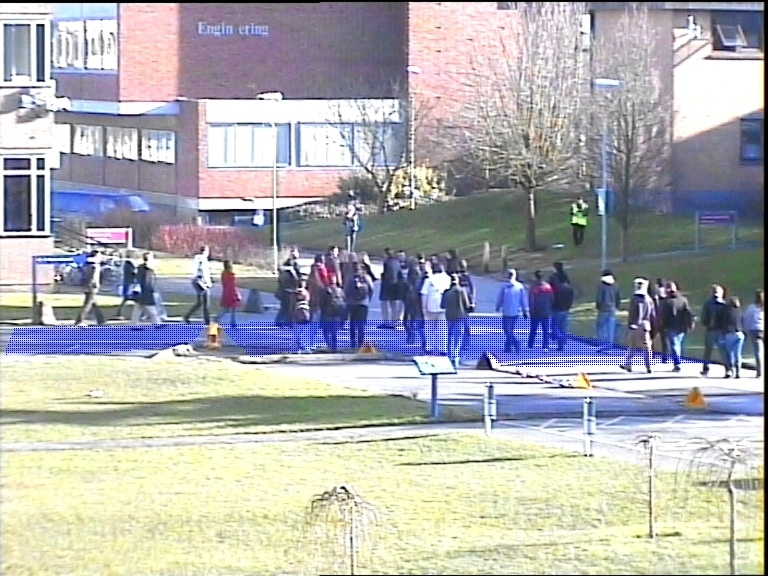}} &\imagetop{\includegraphics[height=0.7\linewidth]{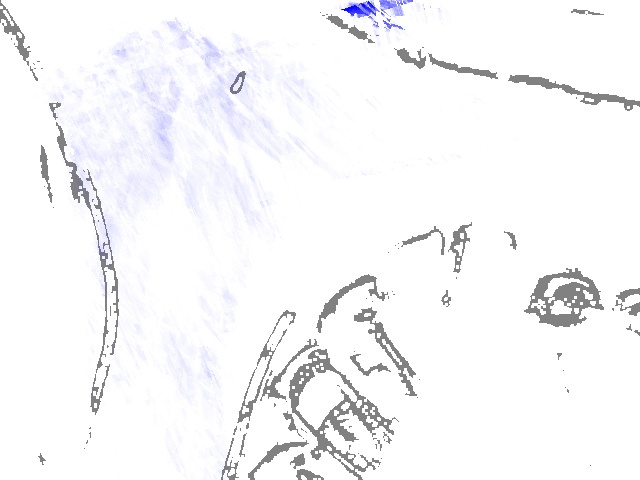}}&\imagetop{\includegraphics[height=0.7\linewidth]{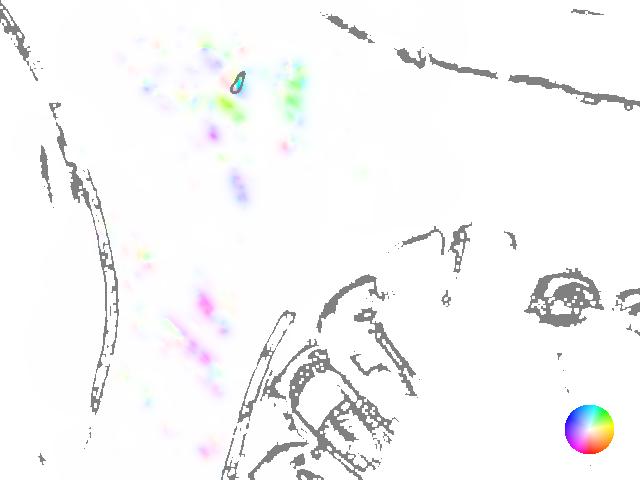}}&Thre flows towards the scene center are recognisable in the cumulative grid and the flow map.\\
\hline
PETS dataset 2009~\cite{pets09}&S3.MF
&The sequence contains a high density crowd, who are performing different activities
. There are overcast lighting conditions
 visible&\imagetop{\includegraphics[height=0.7\linewidth]{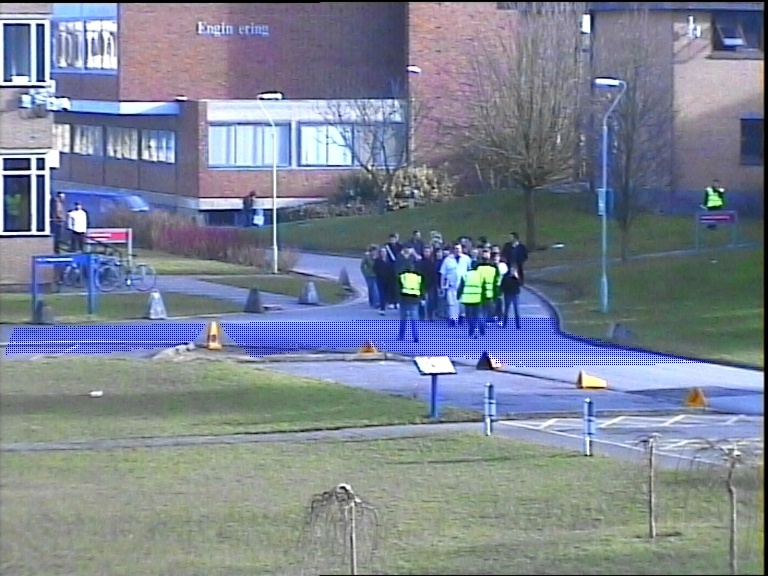}}&\imagetop{\includegraphics[height=0.7\linewidth]{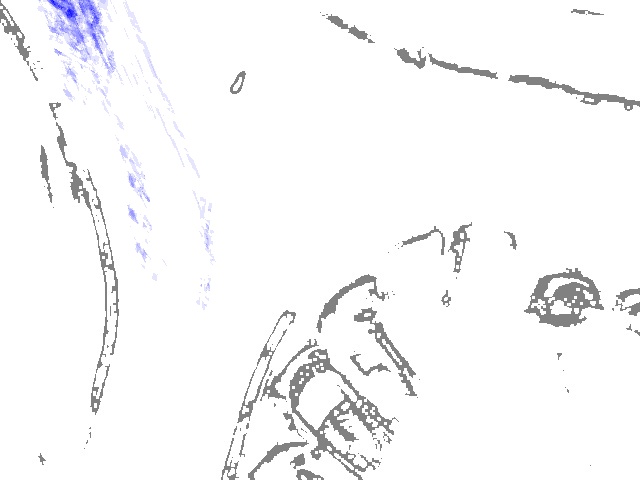}}&\imagetop{\includegraphics[height=0.7\linewidth]{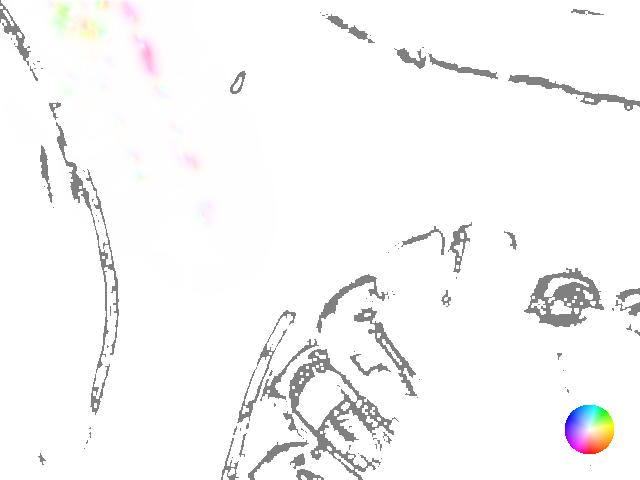}}&Two disjunct streaks from the bottom to the top left of the grid indicate the paths of the men in yellow vest who are approaching the main crowd, which walks towards the scene centre.\\
\hline
\end{tabular}
\end{table*}
\begin{table*}[t]
\caption{The results of our algorithm applied to different sequences of the Depth Dataset~\cite{li2012pedestrian}}
\centering
\label{tab:resultsdepth}      
\begin{tabular}{p{1cm}p{1cm}p{3cm}p{2cm}p{2cm}p{2cm}p{3cm}}
\hline
Depth Dataset \cite{li2012pedestrian}&S1&The sequence contains two people. One leaves the scene, one remains at his position. &\imagetop{\includegraphics[height=0.7\linewidth]{depth01.png}} &\imagetop{\includegraphics[height=0.7\linewidth]{cumulative01.png}}&\imagetop{\includegraphics[height=0.7\linewidth]{flow_data1_04.png}}&The cumulative grid shows the trail of the one person moving from the scene and a blue saturated spot indicates that the other person remains still.\\
\hline
Depth Dataset \cite{li2012pedestrian}&S2&The sequence contains one person who is moving quickly away from the camera &\imagetop{\includegraphics[height=0.7\linewidth]{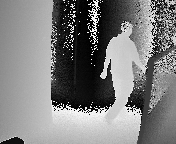}} &\imagetop{\includegraphics[height=0.7\linewidth]{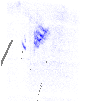}}&\imagetop{\includegraphics[height=0.7\linewidth]{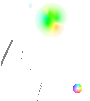}}&The cumulative grid shows the trail of his body positions.\\
\hline
Depth Dataset \cite{li2012pedestrian}&S4&The sequence contains one person who is moving quickly parallel to the image plane &\imagetop{\includegraphics[height=0.7\linewidth]{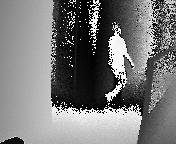}} &\imagetop{\includegraphics[height=0.7\linewidth]{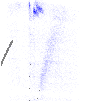}}&\imagetop{\includegraphics[height=0.7\linewidth]{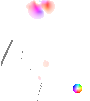}}&The cumulative grid shows the trail of his body positions.\\
\hline
Depth Dataset \cite{li2012pedestrian}&S6&The sequence contains two people. One moves away to the left, the other moves a bit to the camera&\imagetop{\includegraphics[height=0.7\linewidth]{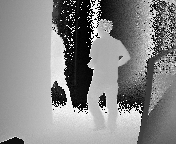}} &\imagetop{\includegraphics[height=0.7\linewidth]{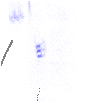}}&\imagetop{\includegraphics[height=0.7\linewidth]{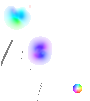}}&The cumulative grid shows the trail of both body positions.\\
\hline
Depth Dataset \cite{li2012pedestrian}&S9&The sequence contains three people. One vanishes to the rear right, one moves parallel to the image plane towards the right and one remains still&\imagetop{\includegraphics[height=0.7\linewidth]{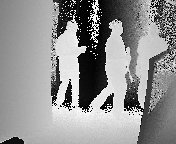}} &\imagetop{\includegraphics[height=0.7\linewidth]{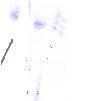}}&\imagetop{\includegraphics[height=0.7\linewidth]{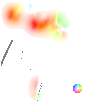}}&The cumulative grid shows the trail of all three body positions.\\
\hline
Depth Dataset \cite{li2012pedestrian}&S13&The sequence contains one person who slowly moves from the rear left towards the scene centre. &\imagetop{\includegraphics[height=0.7\linewidth]{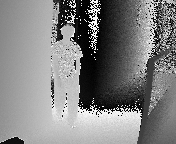}} &\imagetop{\includegraphics[height=0.7\linewidth]{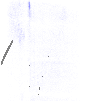}}&\imagetop{\includegraphics[height=0.7\linewidth]{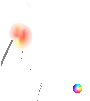}}&The slight fainted trail indicates the motion.\\
\hline
Depth Dataset \cite{li2012pedestrian}&S16&The sequence contains three people. One moves towards the camera, the second moves to the right, while the third remains still at the rear left. &\imagetop{\includegraphics[height=0.7\linewidth]{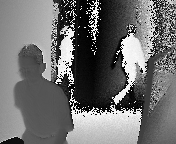}} &\imagetop{\includegraphics[height=0.7\linewidth]{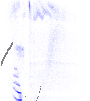}}&\imagetop{\includegraphics[height=0.7\linewidth]{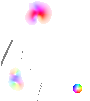}}&The two trails are claerly visible, one spot indicates the person standing still.\\
\hline
Depth Dataset \cite{li2012pedestrian}&S19&The sequence contains one person who quickly moves towards the camera. &\imagetop{\includegraphics[height=0.7\linewidth]{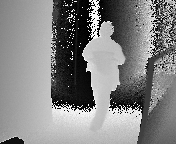}} &\imagetop{\includegraphics[height=0.7\linewidth]{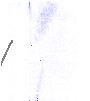}}&\imagetop{\includegraphics[height=0.7\linewidth]{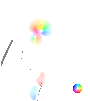}}&The less saturated trail indicates a fast paced motion.\\
\hline
Depth Dataset \cite{li2012pedestrian}&S22&The sequence contains one people who quickly moves from rear left to rear right. &\imagetop{\includegraphics[height=0.7\linewidth]{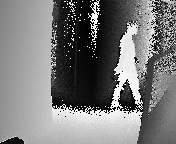}} &\imagetop{\includegraphics[height=0.7\linewidth]{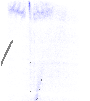}}&\imagetop{\includegraphics[height=0.7\linewidth]{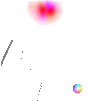}}&The trail is visible in the upper part of the grid.\\
\hline
\end{tabular}
\end{table*}

\begin{table*}[t]
\caption{The results of our algorithm applied to different sequences of the PETS dataset 2007~\cite{pets07}}
\centering
\label{tab:resultspets2007}    
\begin{tabular}{p{1cm}p{1cm}p{3cm}p{2cm}p{2cm}p{2cm}p{3cm}}
\hline
%colorval 45
PETS dataset 2007~\cite{pets07}&S1&The sequence contains one person who enters the scene and then loiters and leaves the scene.&\imagetop{\includegraphics[height=0.7\linewidth]{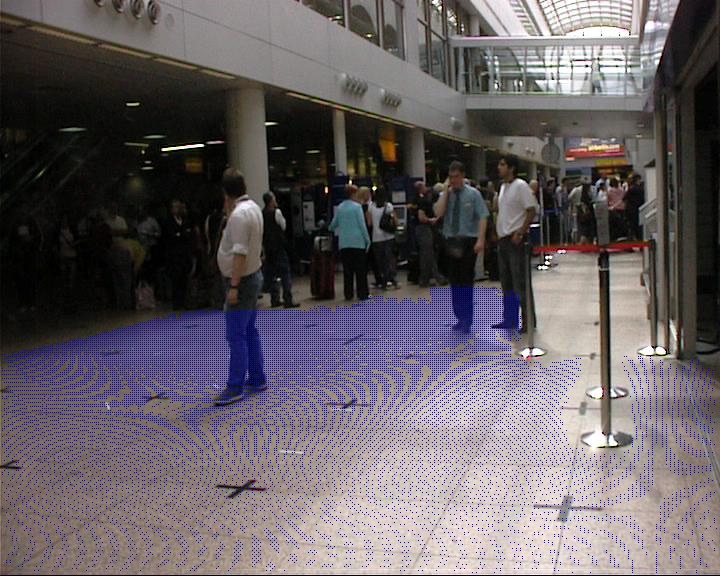}} &\imagetop{\includegraphics[height=0.7\linewidth]{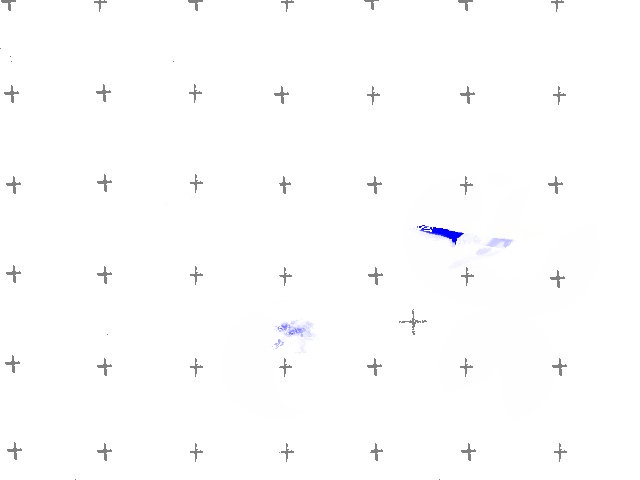}}&\imagetop{\includegraphics[height=0.7\linewidth]{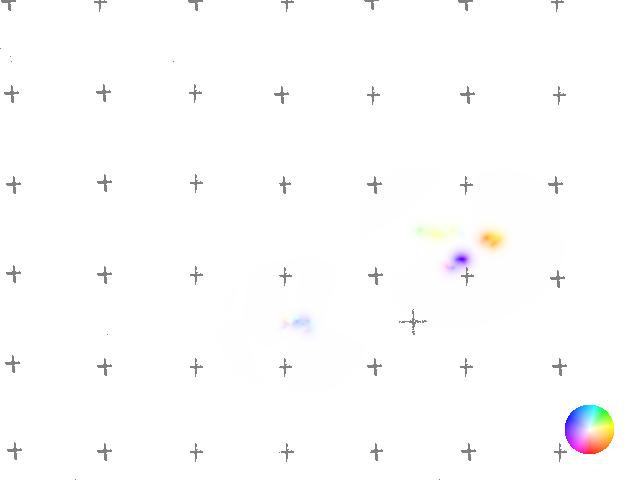}}&In the cumulative grid with moving average $A_t$,$t_{span}=100$ the position of the loitering person emerges visually.\\
\hline
%colorval 85
PETS dataset 2007~\cite{pets07}&S2&The sequence contains a person who walks into the scene and puts a bag on the ground. After loitering the person exits.&\imagetop{\includegraphics[height=0.7\linewidth]{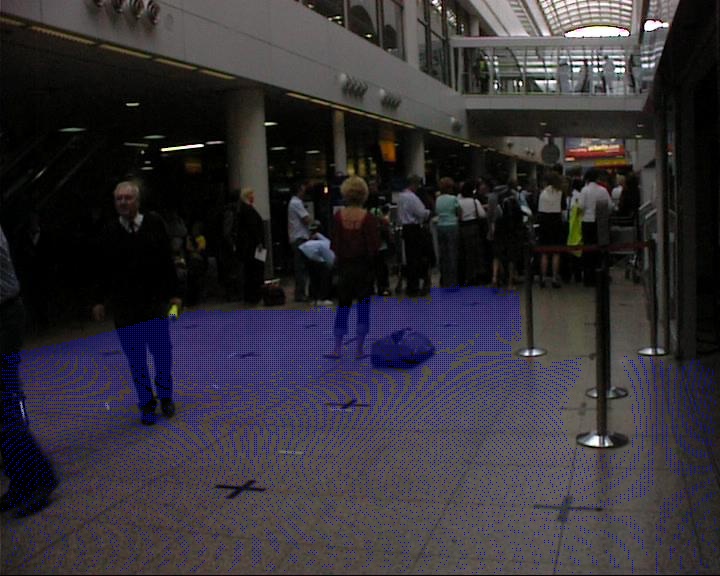}} &\imagetop{\includegraphics[height=0.7\linewidth]{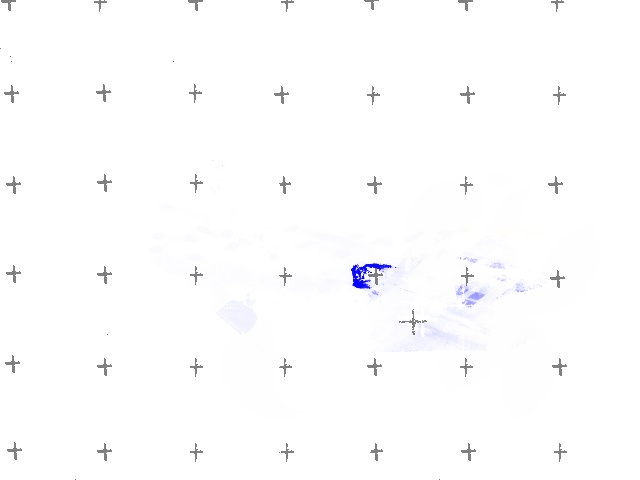}}&\imagetop{\includegraphics[height=0.7\linewidth]{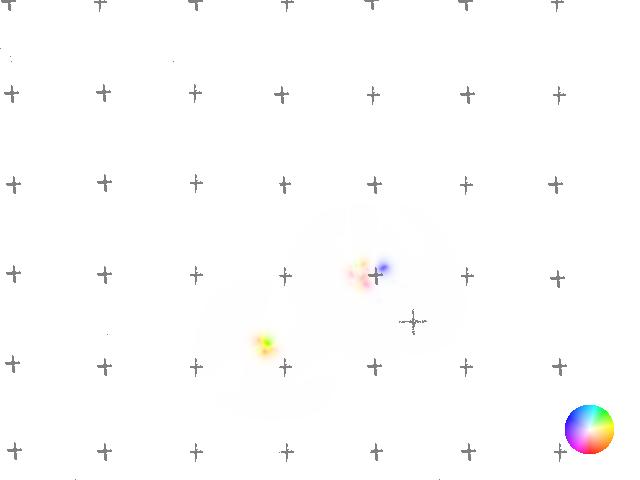}}&The loitering is visible in the cumulative map as blue saturated pixel set. The position of the bag becomes visible in the cumulative map from the moment of positioning until the end of the sequence.\\
\hline
PETS dataset 2007~\cite{pets07}&S3&The sequence contains two people who enter the scene. One places a bag on the ground. The second person picks up the bag and both walk out.&\imagetop{\includegraphics[height=0.7\linewidth]{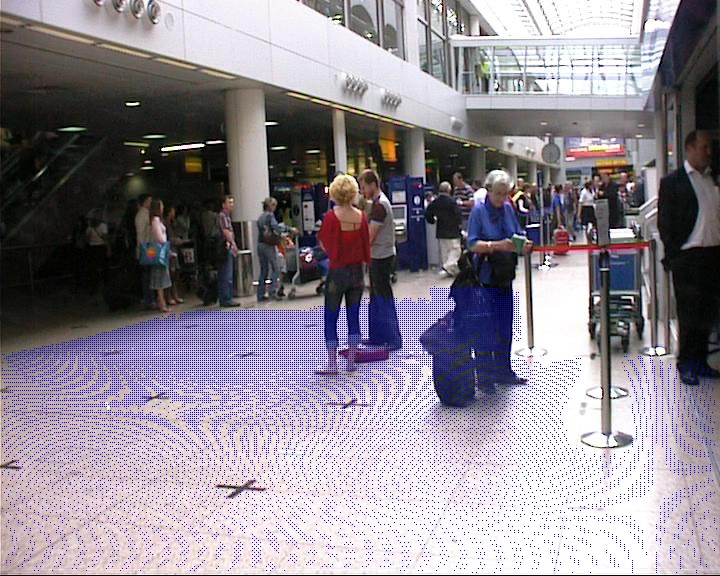}} &\imagetop{\includegraphics[height=0.7\linewidth]{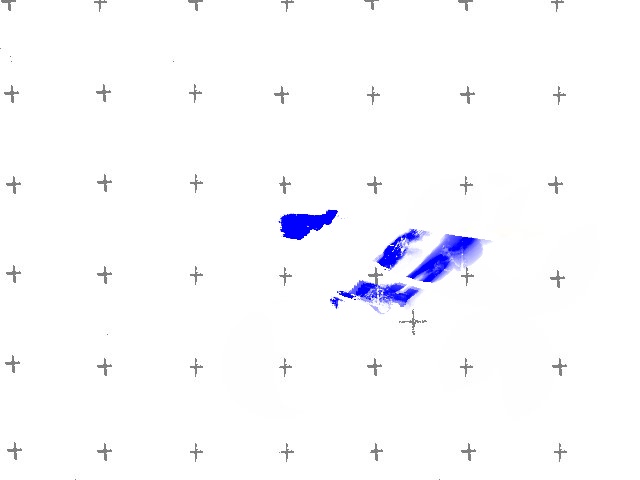}}&\imagetop{\includegraphics[height=0.7\linewidth]{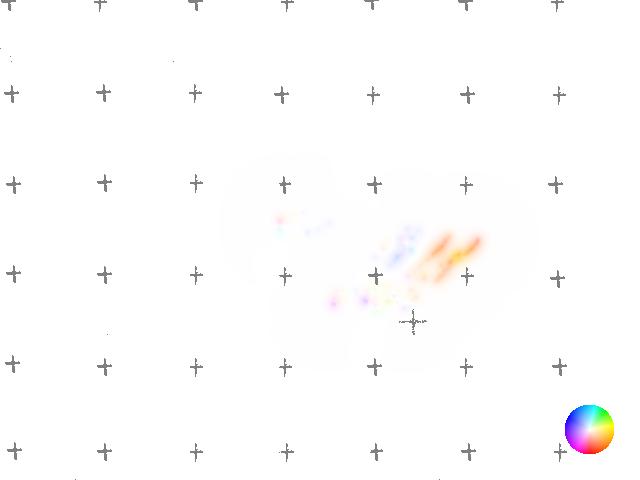}}&Small saturated pixel regions in cumulative grid with moving average $A_t$,$t_{span}=100$ indicate loitering of the two people. The optical flow map shows two spots diverging in different directions.\\
\hline
PETS dataset 2007~\cite{pets07}&S4&The sequence contains four people who walk into the scene. One places a bag to the ground, another picks up the bag and all walk out of the scene.&\imagetop{\includegraphics[height=0.7\linewidth]{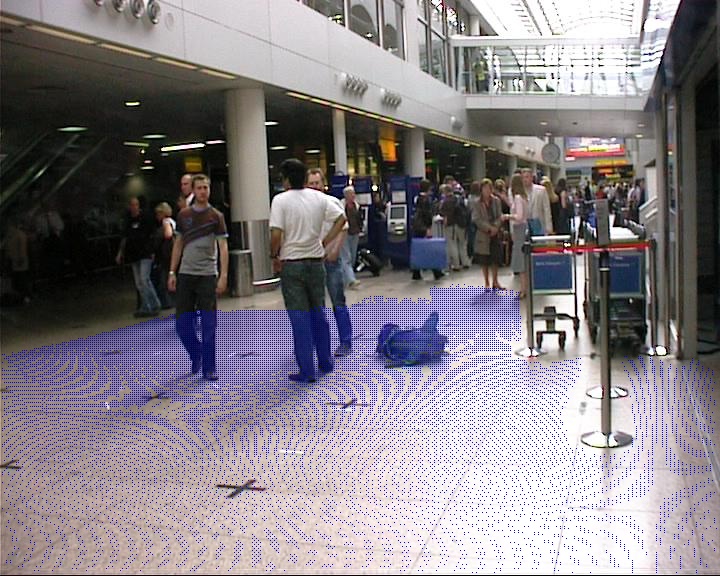}} &\imagetop{\includegraphics[height=0.7\linewidth]{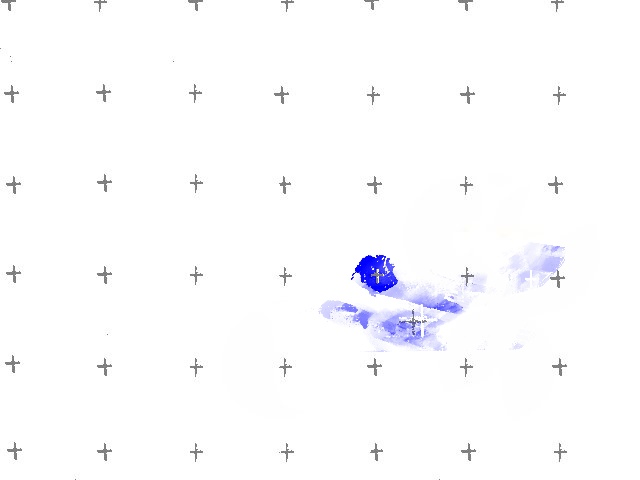}}&\imagetop{\includegraphics[height=0.7\linewidth]{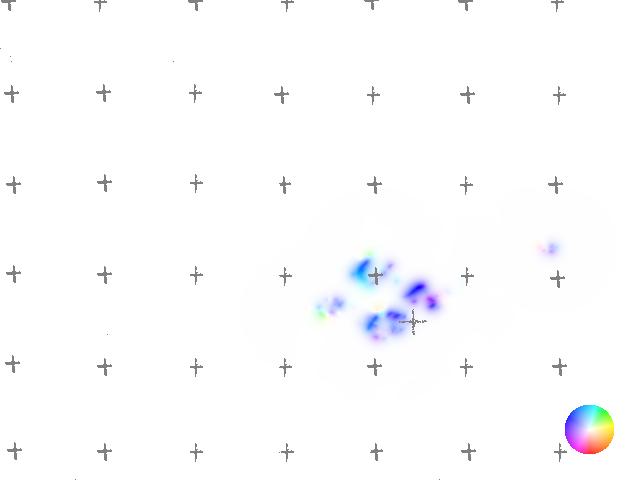}}&The loitering is visible in the cumulative grid with moving average $A_t$,$t_{span}=100$ as blue saturated pixel set.\\
\hline
PETS dataset 2007~\cite{pets07}&S5&The sequence contains one person who the scene. They place the bag on the ground. A second person (thief) picks up the bag and walks out of the scene.&\imagetop{\includegraphics[height=0.7\linewidth]{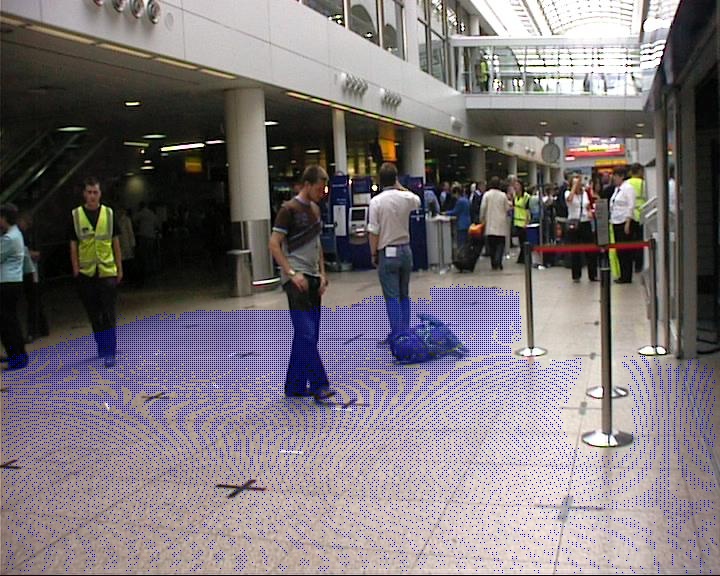}} &\imagetop{\includegraphics[height=0.7\linewidth]{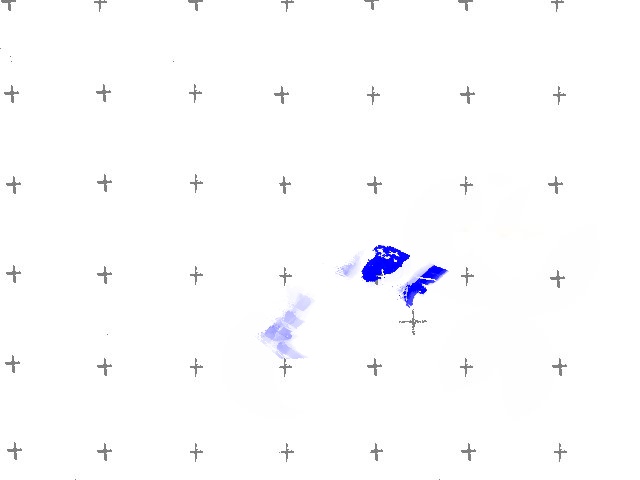}}&\imagetop{\includegraphics[height=0.7\linewidth]{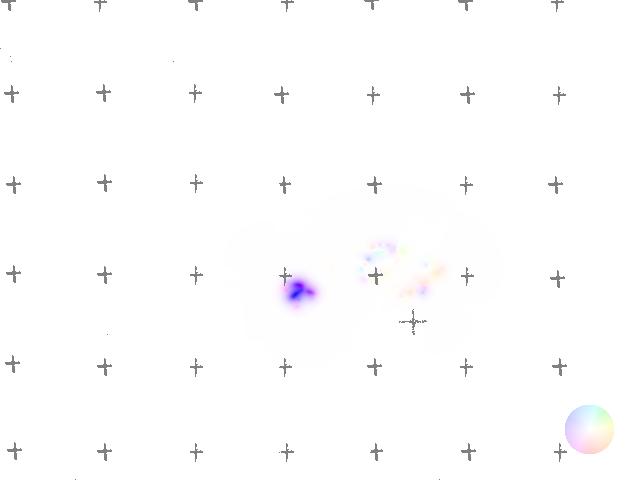}}&In the cumulative grid with moving average $A_t$,$t_{span}=200$, the bag and the owner are visible as blob, with the more saturated pixel set showing the bag. The thiefs approaching motion is seen in a less saturated second blob, that streaks towards the bag.\\
\hline
PETS dataset 2007~\cite{pets07}&S6&The sequence contains two people. They place bags down on the ground. Two other people conduct theft and distraction.&\imagetop{\includegraphics[height=0.7\linewidth]{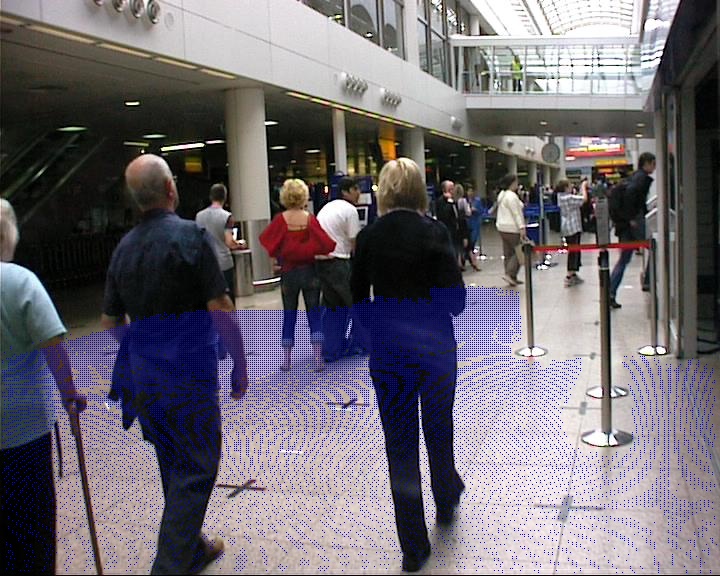}} &\imagetop{\includegraphics[height=0.7\linewidth]{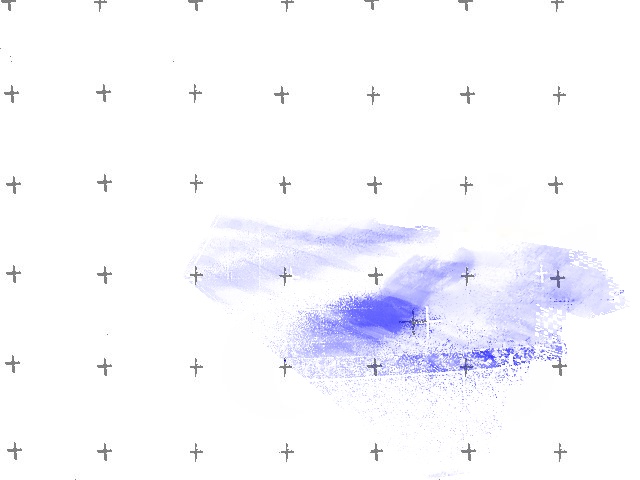}}&\imagetop{\includegraphics[height=0.7\linewidth]{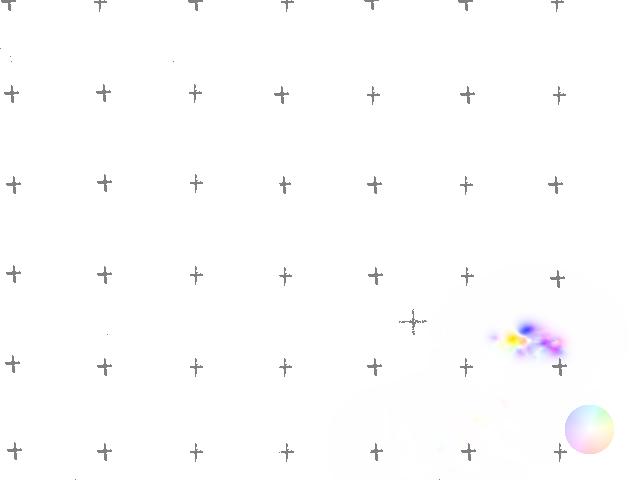}}&Saturated blob in the cumulative grid with moving average $A_t$,$t_{span}=300$ where the two people stand\\
\hline
PETS dataset 2007~\cite{pets07}&S7&The sequence contains a single person with two bags. They place one bag on the ground, leave and return to pick up the left bag.&\imagetop{\includegraphics[height=0.7\linewidth]{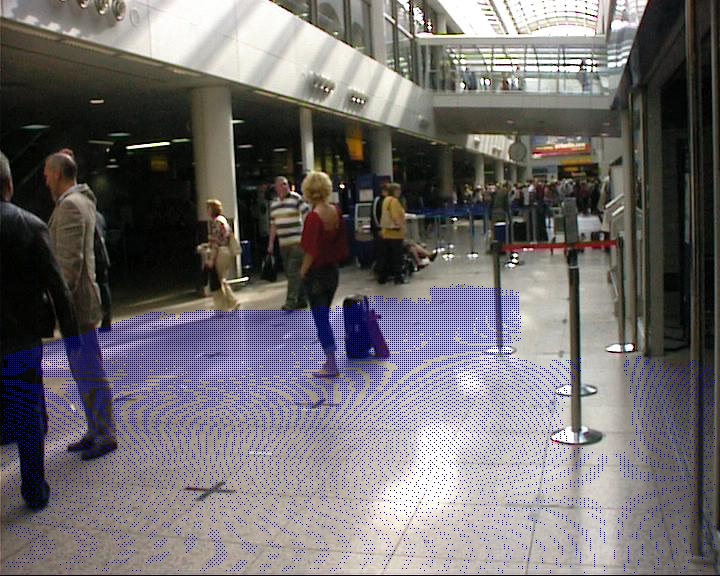}} &\imagetop{\includegraphics[height=0.7\linewidth]{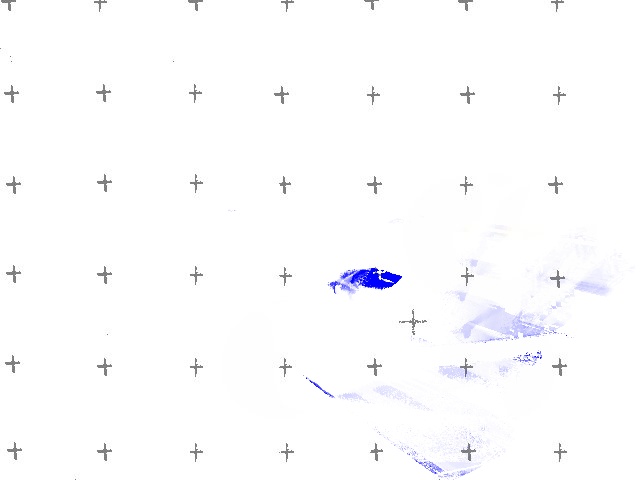}}&\imagetop{\includegraphics[height=0.7\linewidth]{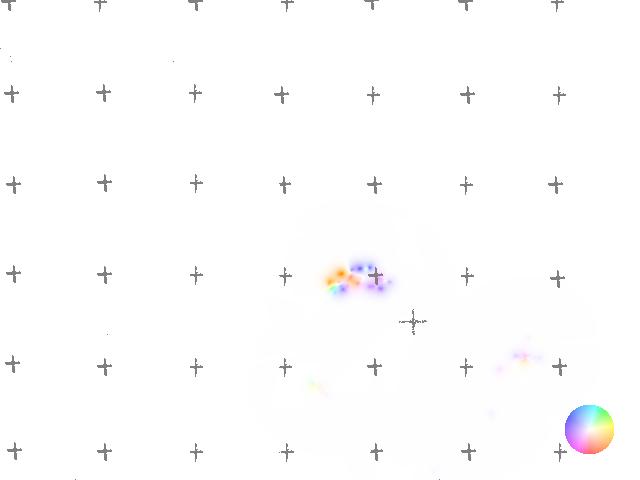}}&Saturated blob in the cumulative grid with moving average $A_t$,$t_{span}=100$ where the person with the luggage stands.\\
\hline
PETS dataset 2007~\cite{pets07}&S8&The sequence contains a person who places a large bag on the ground. He leaves the bag for a short moment before walking away with it .&\imagetop{\includegraphics[height=0.7\linewidth]{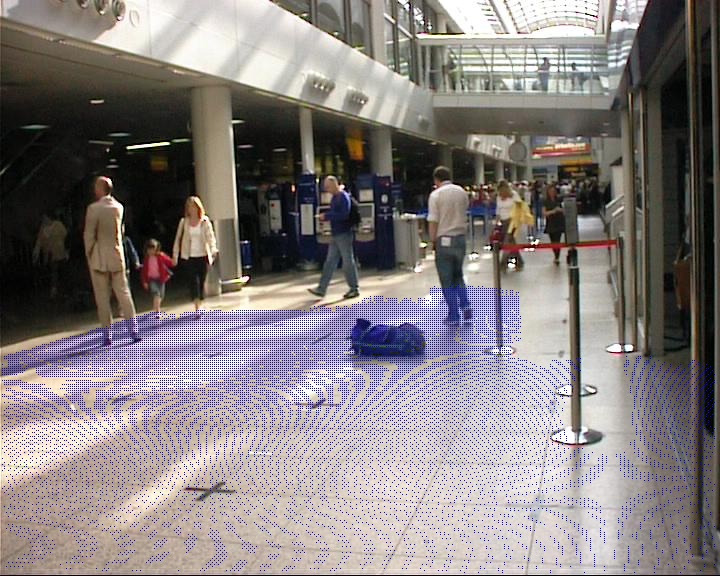}} &\imagetop{\includegraphics[height=0.7\linewidth]{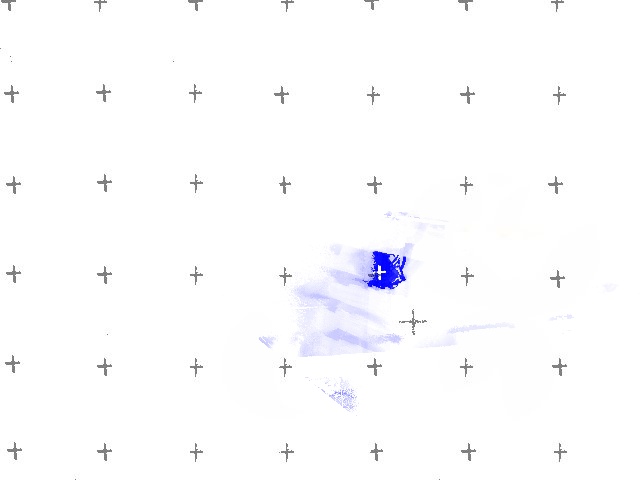}}&\imagetop{\includegraphics[height=0.7\linewidth]{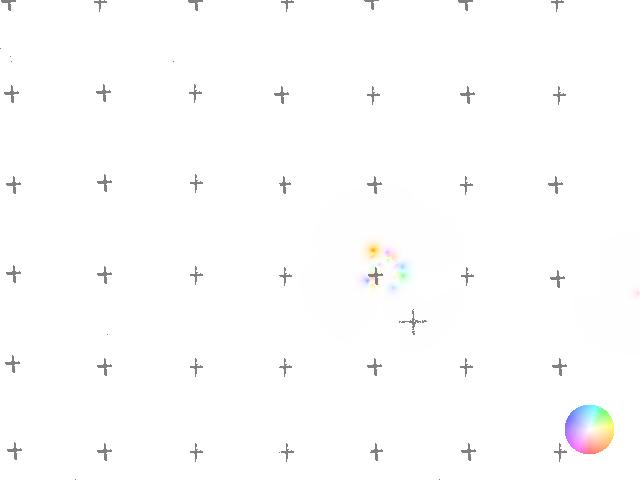}}&Saturated blob in the cumulative grid with moving average $A_t$,$t_{span}=100$ where the abandoned luggage lays.\\
\hline
\end{tabular}
\end{table*}

\section{Conclusion}
\label{conclusion}
We have presented an approach to retarget pedestrian surveillance footage into the topview in order to perform motion and video analysis to it.
Two algorithms, one for RGB-data and one for depth data have been described in detail. The motion analysis consisted of cumulative maps to help crowding detection
and the application of optical flows. We have applied the algorithm to an RGB outdoor dataset consting of eight calibrated cameras and two indoor datasets. One dataset was
a depth-based dataset consisting of one ToF-camera, the other was the surveillance of a train station with for extrinsically calibrated RGB cameras.
\bibliographystyle{plain}
%%use following if all content of bibtex file should be shown
%\nocite{*}
\bibliography{template}

\begin{thebibliography}{10}

\bibitem{pets07}
PETS 2007.
\newblock Dataset of the tenth ieee international workshop on performance
  evaluation of tracking and surveillance.
\newblock \emph{http://www.cvg.rdg.ac.uk/PETS2007/}.

\bibitem{pets09}
PETS 2009.
\newblock Dataset of the eleventh ieee international workshop on performance
  evaluation of tracking and surveillance.
\newblock \emph{http://www.cvg.rdg.ac.uk/PETS2009/}.

\bibitem{article3}
A~Andriyenko, K~Schindler, and S~Roth.
\newblock Discrete-continuous optimization for multi-target tracking.
\newblock {\em CVPR}, 2012.

\bibitem{berger2011markerless}
Kai Berger, Kai Ruhl, Christian Br{\"u}mmer, Yannic Schr{\"o}der, Alexander
  Scholz, and Marcus Magnor.
\newblock Markerless motion capture using multiple color-depth sensors.
\newblock In {\em Proc. Vision, Modeling and Visualization (VMV)}, volume 2011,
  page~3, 2011.

\bibitem{Borgo:2013:CGF}
R.~Borgo., M.~Chen, B.~Daubney, E.~Grundy, H.~Jaenicke, G.~Heidemann,
  B.~Hoeferlin, M.~Hoeferlin, D.~Weiskopf, and X.~Xie.
\newblock A survey on video-based graphics and video visualization.
\newblock In {\em Eurographics 2011 STAR}, 2011.

\bibitem{Botchen:2008:ABM}
Ralf~P. Botchen, Sven Bachthaler, Fabian Schick, Min Chen, Greg Mori, Daniel
  Weiskopf, and Thomas Ertl.
\newblock Action-based multifield video visualization.
\newblock 14(4):885--899, 2008.

\bibitem{butler2012shake}
D~Alex Butler, Shahram Izadi, Otmar Hilliges, David Molyneaux, Steve Hodges,
  and David Kim.
\newblock Shake'n'sense: reducing interference for overlapping structured light
  depth cameras.
\newblock In {\em Proceedings of the 2012 ACM annual conference on Human
  Factors in Computing Systems}, pages 1933--1936. ACM, 2012.

\bibitem{Carroll:2010:IWA:1778765.1778864}
Robert Carroll, Aseem Agarwala, and Maneesh Agrawala.
\newblock Image warps for artistic perspective manipulation.
\newblock {\em ACM Trans. Graph.}, 29(4):127:1--127:9, July 2010.

\bibitem{vis06-chen}
Min Chen, Ralf~P. Botchen, Rudy~R. Hashim, Daniel Weiskopf, Thomas Ertl, and
  Ian~M. Thornton.
\newblock {Visual Signatures in Video Visualization}.
\newblock {\em IEEE Transactions on Visualization and Computer Graphics},
  {12}({5}):1093--1100, 2006.

\bibitem{DalMutto3DPVT10}
Carlo {Dal Mutto}, Pietro Zanuttigh, and Guido~Maria Cortelazzo.
\newblock A probabilistic approach to tof and stereo data fusion.
\newblock In {\em {3DPVT}}, Paris, France, May 2010.

\bibitem{vidvis}
G.W. Daniel and M.~Chen.
\newblock Video visualization.
\newblock {\em Proc. IEEE Visualization}, pages 409--416, October 2003.

\bibitem{article11}
P~Dollar, S~Belongie, and P~Perona.
\newblock The fastest pedestrian detector in the west.
\newblock {\em BMVC}, 2010.

\bibitem{article12}
P~Felzenszwalb, R~Girshick, D~McAllester, and D~Ramanan.
\newblock Object detection with discriminatively trained part-based models.
\newblock {\em IEEE Trans. Pattern Anal. Mach. Intell.}, 2010.

\bibitem{fischer2011combination}
J.~Fischer, G.~Arbeiter, and A.~Verl.
\newblock Combination of time-of-flight depth and stereo using semiglobal
  optimization.
\newblock In {\em Int. Conf. on Robotics and Automation (ICRA)}, pages
  3548--3553. IEEE, 2011.

\bibitem{frati2011using}
Valentino Frati and Domenico Prattichizzo.
\newblock Using kinect for hand tracking and rendering in wearable haptics.
\newblock In {\em World Haptics Conference (WHC), 2011 IEEE}, pages 317--321.
  IEEE, 2011.

\bibitem{girshick2011efficient}
Ross Girshick, Jamie Shotton, Pushmeet Kohli, Antonio Criminisi, and Andrew
  Fitzgibbon.
\newblock Efficient regression of general-activity human poses from depth
  images.
\newblock In {\em Computer Vision (ICCV), 2011 IEEE International Conference
  on}, pages 415--422. IEEE, 2011.

\bibitem{article14}
H~Gong, J~Sim, M~Likhachev, and J~Shi.
\newblock Multi-hypothesis motion planning for visual object tracking.
\newblock {\em ICCV}, 2011.

\bibitem{gudmundsson2008fusion}
S.A. Gudmundsson, H.~Aanaes, and R.~Larsen.
\newblock Fusion of stereo vision and time-of-flight imaging for improved 3d
  estimation.
\newblock {\em IJISTA}, 5(3):425--433, 2008.

\bibitem{hahne2008combining}
U.~Hahne and M.~Alexa.
\newblock Combining time-of-flight depth and stereo images without accurate
  extrinsic calibration.
\newblock {\em IJISTA}, 5(3):325--333, 2008.

\bibitem{hahne2009depth}
U.~Hahne and M.~Alexa.
\newblock Depth imaging by combining time-of-flight and on-demand stereo.
\newblock {\em Dynamic 3D Imaging}, pages 70--83, 2009.

\bibitem{hirschmueller2008}
Heiko Hirschm\"uller.
\newblock Stereo processing by semiglobal matching and mutual information.
\newblock {\em IEEE PAMI}, 30:328--341, 2008.

\bibitem{article17}
C~Huang, B~Wu, and R~Nevatia.
\newblock Robust object tracking by hierarchical association of detection
  responses.
\newblock {\em ECCV}, 2008.

\bibitem{kim2009multi}
Y.M. Kim, C.~Theobalt, J.~Diebel, J.~Kosecka, B.~Miscusik, and S.~Thrun.
\newblock Multi-view image and tof sensor fusion for dense 3d reconstruction.
\newblock In {\em ICCV Workshops}, pages 1542--1549. IEEE, 2009.

\bibitem{kuhnert2006fusion}
K.D. Kuhnert and M.~Stommel.
\newblock Fusion of stereo-camera and pmd-camera data for real-time suited
  precise 3d environment reconstruction.
\newblock In {\em Int. Conf. on Intelligent Robots and Systems}, pages
  4780--4785. IEEE, 2006.

\bibitem{Legg:2011:ICIP}
P.~Legg, M.~Parry, D.~Chung, R.~Jiang, A.~Morris, I.~Griffiths, D.~Marshall,
  and M.~Chen.
\newblock Intelligent filtering by semantic importance for single-view 3d
  reconstruction from snooker video.
\newblock In {\em ICIP}, pages 2433--2436, 2011.

\bibitem{leyvand2011kinect}
Tommer Leyvand, Casey Meekhof, Yi-Chen Wei, Jian Sun, and Baining Guo.
\newblock Kinect identity: Technology and experience.
\newblock {\em Computer}, 44(4):94--96, 2011.

\bibitem{li2012pedestrian}
Yan-Ran Li, Shiqi Yu, and Shengyin Wu.
\newblock Pedestrian detection in depth images using framelet regularization.
\newblock In {\em Computer Science and Automation Engineering (CSAE), 2012 IEEE
  International Conference on}, volume~2, pages 300--303. IEEE, 2012.

\bibitem{maimone2012reducing}
Andrew Maimone and Henry Fuchs.
\newblock Reducing interference between multiple structured light depth sensors
  using motion.
\newblock In {\em Virtual Reality Workshops (VR), 2012 IEEE}, pages 51--54.
  IEEE, 2012.

\bibitem{nair2012high}
Rahul Nair, Frank Lenzen, Stephan Meister, Henrik Sch{\"a}fer, Christoph Garbe,
  and Daniel Kondermann.
\newblock High accuracy tof and stereo sensor fusion at interactive rates.
\newblock In {\em Computer Vision--ECCV 2012. Workshops and Demonstrations},
  pages 1--11. Springer, 2012.

\bibitem{nowozin2011decision}
Sebastian Nowozin, Carsten Rother, Shai Bagon, Toby Sharp, Bangpeng Yao, and
  Pushmeet Kohli.
\newblock Decision tree fields.
\newblock In {\em Computer Vision (ICCV), 2011 IEEE International Conference
  on}, pages 1668--1675. IEEE, 2011.

\bibitem{oikonomidis2011efficient}
Iason Oikonomidis, Nikolaos Kyriazis, and Antonis Argyros.
\newblock Efficient model-based 3d tracking of hand articulations using kinect.
\newblock {\em BMVC, Aug}, 2, 2011.

\bibitem{article21}
T~Parag, F~Porikli, and A~Elgammal.
\newblock Boosting adaptive linear weak classifiers for online learning and
  tracking.
\newblock {\em CVPR}, 2008.

\bibitem{Parry:2011:TVCG}
M.~Parry, P.~Legg, D.~Chung, I.~Griffiths, and M.~Chen.
\newblock Hierarchical event selection for video storyboards with a case study
  on snooker video visualization.
\newblock In {\em IEEE Transactions on Visualization and Computer Graphics},
  pages 1747--1756, 2011.

\bibitem{article15}
H~Pirsiavash, D~Ramanan, and C~Fowlkes.
\newblock Globally-optimal greedy algorithms for tracking a variable number of
  objects.
\newblock {\em CVPR}, 2011.

\bibitem{raheja2011tracking}
Jagdish~L Raheja, Ankit Chaudhary, and Kunal Singal.
\newblock Tracking of fingertips and centers of palm using kinect.
\newblock In {\em Computational Intelligence, Modelling and Simulation
  (CIMSiM), 2011 Third International Conference on}, pages 248--252. IEEE,
  2011.

\bibitem{raptis2011real}
Michalis Raptis, Darko Kirovski, and Hugues Hoppe.
\newblock Real-time classification of dance gestures from skeleton animation.
\newblock In {\em Proceedings of the 2011 ACM SIGGRAPH/Eurographics Symposium
  on Computer Animation}, pages 147--156. ACM, 2011.

\bibitem{Romero:2008:TVCG}
M.~Remero, J.~Summet, J.~Stasko, and G.~Abowd.
\newblock Viz-a-vis: Toward visualizing video through computer vision.
\newblock {\em IEEE Transactions on Visualization and Computer Graphics},
  14(6):1261--1268, 2008.

\bibitem{article19}
M~Rodriguez, I~Laptev, J~Sivic, and J~Audibert.
\newblock Density-aware person detection and tracking in crowds.
\newblock {\em ICCV}, 2011.

\bibitem{sacht2011scalable}
L.~Sacht, L.~Velho, D.~Nehab, and M.~Cicconet.
\newblock Scalable motion-aware panoramic videos.
\newblock In {\em SIGGRAPH Asia 2011 Sketches}, page~37. ACM, 2011.

\bibitem{schroder2011multiple}
Yannic Schr{\"o}der, Alexander Scholz, Kai Berger, Kai Ruhl, Stefan Guthe, and
  Marcus Magnor.
\newblock Multiple kinect studies.
\newblock {\em Computer Graphics}, 2011.

\bibitem{setlur2005automatic}
V.~Setlur, S.~Takagi, R.~Raskar, M.~Gleicher, and B.~Gooch.
\newblock Automatic image retargeting.
\newblock In {\em Proceedings of the 4th international conference on Mobile and
  ubiquitous multimedia}, pages 59--68. ACM, 2005.

\bibitem{article22}
H~Shitrit, J~Berclaz, F~Fleuret, and P.~Fua.
\newblock Tracking multiple people under global appearance constraints.
\newblock {\em ICCV}, 2011.

\bibitem{shotton2011real}
Jamie Shotton, Andrew Fitzgibbon, Mat Cook, Toby Sharp, Mark Finocchio, Richard
  Moore, Alex Kipman, and Andrew Blake.
\newblock Real-time human pose recognition in parts from single depth images.
\newblock In {\em Computer Vision and Pattern Recognition (CVPR), 2011 IEEE
  Conference on}, pages 1297--1304. IEEE, 2011.

\bibitem{article4}
B~Song, T~Jeng, E~Staudt, and A~Roy-Chowdhury.
\newblock A stochastic graph evolution framework for robust multi-target
  tracking.
\newblock {\em ECCV}, 2010.

\bibitem{van2011real}
Michael Van~den Bergh, Daniel Carton, Roderick De~Nijs, Nikos Mitsou, Christian
  Landsiedel, Kolja Kuehnlenz, Dirk Wollherr, Luc Van~Gool, and Martin Buss.
\newblock Real-time 3d hand gesture interaction with a robot for understanding
  directions from humans.
\newblock In {\em RO-MAN, 2011 IEEE}, pages 357--362. IEEE, 2011.

\bibitem{Wang:2007:TVCG}
Yi~Wang, D.~M. Krum, E.~M. Coelho, and D.~A. Bowman.
\newblock Contextualized videos: Combining videos with environment models to
  support situational understanding.
\newblock {\em IEEE Transactions on Visualization and Computer Graphics},
  13(6):1568--1575, 2007.

\bibitem{article20}
M~Yang, Y~Wu, and G~Hua.
\newblock Context-aware visual tracking.
\newblock {\em IEEE Trans. Pattern Anal. Mach. Intell.}, 2009.

\bibitem{mmsp-10-qingxiong-yang}
Q.~Yang, K.-H. Tan, B.~Culbertson, and J.~Apostolopoulos.
\newblock Fusion of active and passive sensors for fast 3d capture.
\newblock In {\em MMSP}, 2010.

\bibitem{article25}
J~Yao and J~Odobez.
\newblock Multi-camera multi-person 3d space tracking with mcmc in surveillance
  scenarios.
\newblock {\em ECCV, M2SFA2 Workshop}, 2008.

\bibitem{zollhofer2011automatic}
Michael Zollh{\"o}fer, Michael Martinek, G{\"u}nther Greiner, Marc Stamminger,
  and Jochen S{\"u}{\ss}muth.
\newblock Automatic reconstruction of personalized avatars from 3d face scans.
\newblock {\em Computer Animation and Virtual Worlds}, 22(2-3):195--202, 2011.

\end{thebibliography}

\end{document}